\documentstyle[aclap,calc,ifthen,parsetree]{article}

\author{George Anton Kiraz%
\thanks{\hspace{.1in}Supported by a Benefactor Studentship
from St John's College. This research was done under the supervision of
Dr Stephen G. Pulman. Thanks to the anonymous reviewers for their
comments. All mistakes remain mine.}\\
Computer Laboratory \\ 
University of Cambridge (St John's College)\\
Email: {\tt George.Kiraz@cl.cam.ac.uk}\\
URL: {\tt http://www.cl.cam.ac.uk/users/gk105}}

\title{\vspace{-0.5in}\d{S}EM\d{H}E: A Generalised Two-Level System}

\begin{document}
\maketitle
\bibliographystyle{fullname}
\vspace{-0.5in}


\newcommand{\Syl}{$\sigma$}             
\newcommand{\Mor}{$\mu$}                
\newcommand{\Sylm}{$\sigma_{\mu}$}      
\newcommand{\Sylmm}{$\sigma_{\mu\mu}$}  
\newcommand{\Sylx}{$\sigma_{x}$} 
\newcommand{\kernel}{B:$\Phi$}          
\newcommand{\residue}{B/$\Phi$}         
\newcommand{\ppc}{O:$\Phi$}             
\newcommand{\npc}{O/$\Phi$}             

\newcommand{\composition}{$\circ$}      
\newcommand{\conc}{$^{\frown}$}         
\newcommand{\estr}{$\varepsilon$}   
\newcommand{\func}[2]
   {\mbox{{\sc #1(}#2{\sc )}}}

\newcommand{\lab}{$\langle$} 
\newcommand{\rab}{$\rangle$} 


\newcounter{boxwidth}
\newcounter{boxhight}
\newcounter{mttlmboxwidth}
\newcounter{mttlmboxhight}
\newcounter{autosegboxwidth}
\newcounter{defaulthight}
\newcounter{strlength}
\newcounter{notiers}
\newcounter{picwidth}
\newcounter{pichight}
\newcounter{lift}
\newcounter{linelen}
\newcounter{xdirection}
\newcounter{ydirection}
\newcounter{curx}
\newcounter{cury}
\newcounter{pictopmargin}

\newlength{\tapenamewidth}
\newlength{\templen}

\setcounter{mttlmboxwidth}{12}%
\setcounter{mttlmboxhight}{12}
\setcounter{autosegboxwidth}{10}%
\setcounter{pictopmargin}{2}

\newcounter{fsmboxwidth}  \setcounter{fsmboxwidth}{30}
\newcounter{fsmcolumns}  \setcounter{fsmcolumns}{7}
\newcounter{fsmrows}
\newcounter{fsmcolumnsx}
\newcounter{fsmrowsx}
\newcounter{nostates}
\newcounter{curstate}
\newcounter{fsmradius}

\newcommand{\fsm}[2]
   {
    \setcounter{boxwidth}{\value{fsmboxwidth}}%
    \setcounter{boxhight}{\value{fsmboxwidth}}%
    \setcounter{nostates}{0}%
    \countstates#1|| END%
    \setcounter{fsmrows}{\value{nostates}/\value{fsmcolumns}}%
    \setcounter{temp}{\value{fsmrows}*\value{fsmcolumns}}%
    \ifthenelse{\value{nostates} = \value{temp}}%
               {}%
               {\stepcounter{fsmrows}}%
    \setcounter{fsmcolumnsx}{\value{fsmcolumns}-1}%
    \setcounter{fsmrowsx}{\value{fsmrows}-1}%
    \setcounter{fsmradius}{\value{boxwidth}-\value{boxwidth}/10}%
    \setcounter{picwidth}{\value{boxwidth}*\value{fsmcolumns}+%
                          \value{boxwidth}*\value{fsmcolumnsx}+%
                          \value{boxwidth}*2}%
    \setcounter{pichight}{\value{boxhight}*\value{fsmrows}+%
                          \value{boxhight}*\value{fsmrowsx}+%
                          \value{boxhight}*2}%
    \setcounter{lift}{\value{pichight}/-2}%
    \rule[\the\value{lift}pt]{3 pt}{\the\value{pichight}pt}%
    \begin{picture}(\the\value{picwidth},0)(0,-\the\value{lift})%
       \setcounter{curstate}{0}%
       \setcounter{curx}{\value{boxhight}}%
       \setcounter{cury}{\value{boxhight}*\value{fsmrowsx}*2+\value{boxhight}}%
       \drawstates#1|| END%
    \end{picture}
   }

\def\drawstates#1,#2,#3|#4 END
   {
    \put(\the\value{curx},\the\value{cury})%
         {\makebox(\the\value{boxwidth},\the\value{boxhight}){#1}}%
    \setcounter{curx}{\value{curx}+\value{boxwidth}/2}%
    \setcounter{cury}{\value{cury}+\value{boxhight}/2}%
    \put(\the\value{curx},\the\value{cury})%
         {\circle{\the\value{boxwidth}}}%
    \ifthenelse{\equal{#3}{y}}%
       {\put(\the\value{curx},\the\value{cury})%
            {\circle{\the\value{fsmradius}}}}%
       {}%
    \setcounter{curx}{\value{curx}-\value{boxwidth}/2}%
    \setcounter{cury}{\value{cury}-\value{boxhight}/2}%
    \stepcounter{curstate}%
    \ifthenelse{\value{curstate} = \value{fsmcolumns}}
               {\setcounter{curx}{\value{boxhight}}%
                \setcounter{cury}{\value{cury}-\value{boxhight}*2}%
                \setcounter{curstate}{0}}%
               {\setcounter{curx}{\value{curx}+\value{boxwidth}*2}}%
    \ifthenelse{\equal{#4}{|}}{}%
               {\drawstates#4 END}%
   }

\def\countstates#1|#2 END
   {\stepcounter{nostates}%
    \ifthenelse{\equal{#2}{|}}{}%
               {\countstates#2 END}%
   }


\newcommand{\mttlmsetwidth}[1]%
   {\setcounter{mttlmboxwidth}{#1}}

\newcommand{\mttlmsethight}[1]%
   {\setcounter{mttlmboxhight}{#1}}

\newcommand{\mttlm}[3]
   {\immediate\write16{MTTLM = #1}%
    \setcounter{boxwidth}{\value{mttlmboxwidth}}%
    \setcounter{boxhight}{\value{mttlmboxhight}}%
    \setcounter{strlength}{0}%
    \setcounter{notiers}{0}%
    \setlength{\tapenamewidth}{0pt}%
    \counttiers#3|| END%
    \countsurfacewidth#1 END%
    \findtapenamewidth#3|| END%
    \setcounter{picwidth}{\value{boxwidth}*\value{strlength}}%
    \setcounter{pichight}{\value{boxhight}*\value{notiers}+%
                          2*\value{boxhight}+\value{pictopmargin}}%
    \setcounter{lift}{\value{pichight}/-2}%
    \rule[\the\value{lift}pt]{0 pt}{\the\value{pichight}pt}%
    \begin{picture}(\the\value{picwidth},0)(0,-\the\value{lift})%
       \let\boxtype=\makebox%
       \setcounter{curx}{0}%
       \setcounter{cury}{0}%
       \displayonetape#1 END%
       \setcounter{curx}{0}%
       \setcounter{cury}{\value{boxhight}}%
       \displaystrings#2|| END%
       \setcounter{curx}{0}%
       \setcounter{cury}{2*\value{boxhight}}%
       \displaymanytapes#3|| END%
    \end{picture}%
    \hspace{\the\tapenamewidth}%
   }

\def\countsurfacewidth#1:#2 END%
   {\countwidth#1| END%
    \maxtapenamewidth#2 END%
   }

\def\displayonetape#1:#2 END%
   {
    \put(\the\value{curx},\the\value{cury})%
         {\framebox(\the\value{picwidth},\the\value{boxhight}){}}%
    \partition#1| END%
    \put(\the\value{curx},\the\value{cury})%
         {\makebox(\the\value{boxwidth},\the\value{boxhight})[l]{\ {\em #2}}}%
    \setcounter{curx}{0}%
    \displaystrings#1|| END%
   }    

\def\displaymanytapes#1|#2 END%
   {\displayonetape#1 END%
    \addtocounter{cury}{-\value{boxhight}}%
    \ifthenelse{\equal{#2}{|}}{}%
               {\displaymanytapes#2 END}%
   }

\def\partition#1#2 END%
   {\addtocounter{curx}{\value{boxwidth}}%
    \ifthenelse{\equal{#1}{-}}{}%
                      {\put(\value{curx},\value{cury})%
                         {\dashbox{0.5}(0,\value{boxhight}){}}}
    \ifthenelse{\equal{#2}{|}}{}%
               {\partition#2 END}%
   }

\def\findtapenamewidth#1:#2|#3 END
   {\maxtapenamewidth#2 END%
    \ifthenelse{\equal{#3}{|}}{}%
               {\findtapenamewidth#3 END}%
   }

\def\maxtapenamewidth#1 END%
   {\settowidth{\templen}{{\em #1}}%
    \ifthenelse{\templen > \tapenamewidth}%
               {\settowidth{\tapenamewidth}{\ {\em #1}}}%
               {}%
   }


\newcommand{\autosegsetwidth}[1]%
   {\setcounter{autosegboxwidth}{#1}}

\newcommand{\autoseg}[2]%
   {\immediate\write16{Autseg Tiers #1}%
    \setcounter{boxwidth}{\value{autosegboxwidth}}%
    \setcounter{boxhight}{\value{autosegboxwidth}}%
    \setcounter{strlength}{0}%
    \setcounter{notiers}{0}%
    \counttiers#1|| END%
    \countwidth#1 END%
    \setcounter{picwidth}{\value{boxwidth}*\value{strlength}}%
    \setcounter{pichight}{2*\value{boxwidth}*\value{notiers}-%
                          \value{boxwidth}+\value{pictopmargin}}%
    \setcounter{lift}{\value{pichight}/-2}%
    \rule[\the\value{lift}pt]{0 pt}{\the\value{pichight}pt}%
    \begin{picture}(\the\value{picwidth},0)(0,-\the\value{lift})%
       \let\boxtype=\makebox%
       \setcounter{curx}{0}%
       \setcounter{cury}{0}%
       \displaystrings#1|| END%
       \setcounter{curx}{\value{boxwidth}/2}%
       \linkstrings#2,||| END%
    \end{picture}%
   }

\def\counttiers#1|#2 END
   {\stepcounter{notiers}%
    \ifthenelse{\equal{#2}{|}}{}%
               {\counttiers#2 END}%
   }

\def\countwidth#1|#2 END%
   {\countlength#1| END%
   }

\def\countlength#1#2 END
   {\stepcounter{strlength}%
    \ifthenelse{\equal{#2}{|}}{}%
               {\countlength#2 END}%
   }

\def\displaystrings#1|#2 END
   {\setmorpheme#1| END%
    \setcounter{curx}{0}%
    \addtocounter{cury}{2*\value{boxhight}}%
    \ifthenelse{\equal{#2}{|}}{}%
               {\displaystrings#2 END}%
   }

\def\setmorpheme#1#2 END%
   {\ifthenelse{\equal{#1}{-}}{}%
               {\put(\the\value{curx},\the\value{cury})%
                {\boxtype(\the\value{boxwidth},\the\value{boxhight}){#1}}}%
    \addtocounter{curx}{\value{boxwidth}}%
    \ifthenelse{\equal{#2}{|}}{}%
               {\setmorpheme#2 END}%
   }

\def\linkstrings#1#2#3,#4 END
   {
    \ifthenelse{\equal{#1}{-}}{}%
       {
        \ifthenelse{#1 < #2}%
               {\setcounter{cury}{(2*#1-1)*\value{boxwidth}}%
                \setcounter{ydirection}{1}%
                \setcounter{linelen}{(2*(#2-#1)-1)*\value{boxwidth}}}%
               {}%
        \ifthenelse{#1 > #2}%
               {\setcounter{cury}{2*(#1-1)*\value{boxwidth}}%
                \setcounter{ydirection}{-1}%
                \setcounter{linelen}{(2*(#1-#2)-1)*\value{boxwidth}}}%
               {}%
        \setcounter{xdirection}{0}%
        \ifthenelse{\equal{#3}{r}}{\setcounter{xdirection}{1}}{}%
        \ifthenelse{\equal{#3}{rr}}{\setcounter{xdirection}{2}%
                                   \addtocounter{linelen}{\value{boxwidth}}}{}%
        \ifthenelse{\equal{#3}{rrr}}{\setcounter{xdirection}{3}%
                                   \addtocounter{linelen}{2*\value{boxwidth}}}%
                                   {}%
        \ifthenelse{\equal{#3}{l}}{\setcounter{xdirection}{-1}}{}%
        \ifthenelse{\equal{#3}{ll}}{\setcounter{xdirection}{-2}%
                                   \addtocounter{linelen}{\value{boxwidth}}}{}%
        \ifthenelse{\equal{#3}{lll}}{\setcounter{xdirection}{-3}%
                                   \addtocounter{linelen}{2*\value{boxwidth}}}%
                                   {}%
        \put(\value{curx},\value{cury})%
            {\line(\value{xdirection},\value{ydirection})%
            {\value{linelen}}}}%
    \addtocounter{curx}{\value{boxwidth}}%
    \ifthenelse{\equal{#4}{|||}}{}%
               {\linkstrings#4 END}%
   }

\newcounter{moraboxwidth}
\setcounter{moraboxwidth}{10}

\newcommand{\morasetwidth}[1]%
   {\setcounter{moraboxwidth}{#1}}

\newcommand{\moratree}[2]%
   {
    \setcounter{picwidth}{2*\value{moraboxwidth}}%
    \setcounter{pichight}{4*\value{moraboxwidth}+\value{moraboxwidth}/2}%
    \setcounter{lift}{\value{pichight}/-2}%
    \rule[\the\value{lift}pt]{0 pt}{\the\value{pichight}pt}%
    \begin{picture}(\the\value{picwidth},0)(0,-\the\value{lift})%
       \drawbasic{#1}{#2}%
    \end{picture}%
   }

\newcommand{\mmoratree}[3]%
   {
    \setcounter{picwidth}{3*\value{moraboxwidth}}%
    \setcounter{pichight}{4*\value{moraboxwidth}+\value{moraboxwidth}/2}%
    \setcounter{lift}{\value{pichight}/-2}%
    \rule[\the\value{lift}pt]{0 pt}{\the\value{pichight}pt}%
    \begin{picture}(\the\value{picwidth},0)(0,-\the\value{lift})%
       \drawbasic{#1}{#2}%
       \setcounter{curx}{2*\value{moraboxwidth}}%
       \put(\value{curx},0)%
          {\makebox(\value{moraboxwidth},\value{moraboxwidth})[b]{#3}}%
       \setcounter{cury}{2*\value{moraboxwidth}}%
       \put(\value{curx},\value{cury})%
          {\makebox(\value{moraboxwidth},\value{moraboxwidth}){\Mor}}%
       \setcounter{curx}{2*\value{moraboxwidth}+\value{moraboxwidth}/2}%
       \put(\value{curx},\value{cury}){\line(0,-1){\value{moraboxwidth}}}%
       \setcounter{curx}{\value{moraboxwidth}+\value{moraboxwidth}/2}%
       \setcounter{cury}{3*\value{moraboxwidth}+\value{moraboxwidth}/2}%
       \put(\value{curx},\value{cury}){\line(2,-1){\value{moraboxwidth}}}%
    \end{picture}%
   }

\newcommand{\xmoratree}[1]%
   {
    \setcounter{picwidth}{\value{moraboxwidth}}%
    \setcounter{pichight}{4*\value{moraboxwidth}+\value{moraboxwidth}/2}%
    \setcounter{lift}{\value{pichight}/-2}%
    \rule[\the\value{lift}pt]{0 pt}{\the\value{pichight}pt}%
    \begin{picture}(\the\value{picwidth},0)(0,-\the\value{lift})%
       \put(0,0){\makebox(\value{moraboxwidth},\value{moraboxwidth})[b]{#1}}%
       \setcounter{cury}{3*\value{moraboxwidth}+\value{moraboxwidth}/2}%
       \put(0,\value{cury})%
       {\makebox(\value{moraboxwidth},\value{moraboxwidth}){\Sylx}}%
       \setcounter{curx}{\value{moraboxwidth}/2}%
       \setcounter{linelen}{2*\value{moraboxwidth}+\value{moraboxwidth}/2}%
       \put(\value{curx},\value{cury}){\line(0,-1){\value{linelen}}}%
    \end{picture}%
   }

\newcommand{\gmoratree}[5]%
   {\mbox{\mmoratree{#1}{#2}{#3}%
          \hspace{-\value{moraboxwidth}pt}%
          \mmoratree{{}}{#4}{#5}}%
   \immediate\write16{(#1,#2,#3,#4,#5)}%
   }

\newcommand{\drawbasic}[2]%
   {
    \put(0,0){\makebox(\value{moraboxwidth},\value{moraboxwidth})[b]{#1}}%
    \put(\value{moraboxwidth},0)%
       {\makebox(\value{moraboxwidth},\value{moraboxwidth})[b]{#2}}%
    \setcounter{cury}{2*\value{moraboxwidth}}%
    \put(\value{moraboxwidth},\value{cury})%
       {\makebox(\value{moraboxwidth},\value{moraboxwidth}){\Mor}}%
    \setcounter{cury}{3*\value{moraboxwidth}+\value{moraboxwidth}/2}%
    \put(\value{moraboxwidth},\value{cury})%
       {\makebox(\value{moraboxwidth},\value{moraboxwidth}){\Syl}}%
    \setcounter{curx}{\value{moraboxwidth}+\value{moraboxwidth}/2}%
    \setcounter{cury}{3*\value{moraboxwidth}+\value{moraboxwidth}/2}%
    \put(\value{curx},\value{cury}){\line(-2,-5){\value{moraboxwidth}}}%
    \setcounter{linelen}{\value{moraboxwidth}/2}%
    \put(\value{curx},\value{cury}){\line(0,-1){\value{linelen}}}%
    \setcounter{cury}{2*\value{moraboxwidth}}%
    \put(\value{curx},\value{cury}){\line(0,-1){\value{moraboxwidth}}}%
   }

\newcounter{tapeboxhight}
\setcounter{tapeboxhight}{15}
\newcounter{delta}
\newcounter{fstwidth}
\newcounter{temp}

\newcommand{\tapehight}[1]%
   {\setcounter{tapeboxhight}{#1}}

\newcommand{\cascadetransducers}%
   {
    \setcounter{picwidth}{10*\value{tapeboxhight}}%
    \setcounter{pichight}{9*\value{tapeboxhight}}%
    \setcounter{lift}{\value{pichight}/-2}%
    \rule[\the\value{lift}pt]{0 pt}{\the\value{pichight}pt}%
    \begin{picture}(\the\value{picwidth},0)(0,-\the\value{lift})%
       \put(0,0){\framebox(\value{picwidth},\value{tapeboxhight})
             {Surface String}}%
       \setcounter{cury}{4*\value{tapeboxhight}}
       \setcounter{temp}{8*\value{tapeboxhight}}
       \put(\value{tapeboxhight},\value{cury})
             {\framebox(\value{temp},\value{tapeboxhight})
             {Intermediate String}}%
       \setcounter{cury}{8*\value{tapeboxhight}}
       \put(0,\value{cury}){\framebox(\value{picwidth},\value{tapeboxhight})
             {Lexical String}}%
       \setcounter{curx}{\value{picwidth}/2}
       \setcounter{cury}{\value{tapeboxhight}}
       \setcounter{delta}{2*\value{tapeboxhight}}
       \multiput(\value{curx},\value{cury})(0,\value{delta}){4}
          {\line(0,1){\value{tapeboxhight}}}
       \setcounter{cury}{2*\value{tapeboxhight}+\value{tapeboxhight}/2}
       \setcounter{delta}{4*\value{tapeboxhight}}
       \setcounter{fstwidth}{2*\value{tapeboxhight}}
       \multiput(\value{curx},\value{cury})(0,\value{delta}){2}
          {\oval(\value{fstwidth},\value{tapeboxhight})}
       \put(\value{curx},\value{cury}){\makebox(0,0){\em FST$_n$}}%
       \setcounter{cury}{6*\value{tapeboxhight}+\value{tapeboxhight}/2}
       \put(\value{curx},\value{cury}){\makebox(0,0){\em FST$_1$}}%
    \end{picture}%
    ~$\Longrightarrow$~%
    \begin{picture}(\the\value{picwidth},0)(0,-\the\value{lift})%
       \put(0,0){\framebox(\value{picwidth},\value{tapeboxhight})
             {Surface String}}%
       \setcounter{cury}{8*\value{tapeboxhight}}
       \put(0,\value{cury}){\framebox(\value{picwidth},\value{tapeboxhight})
             {Lexical String}}%
       \setcounter{curx}{\value{picwidth}/2}
       \setcounter{cury}{\value{tapeboxhight}}
       \setcounter{temp}{3*\value{tapeboxhight}}
       \setcounter{delta}{4*\value{tapeboxhight}}
       \multiput(\value{curx},\value{cury})(0,\value{delta}){2}
          {\line(0,1){\value{temp}}}
       \setcounter{cury}{4*\value{tapeboxhight}+\value{tapeboxhight}/2}
       \setcounter{fstwidth}{8*\value{tapeboxhight}}
       \put(\value{curx},\value{cury})
              {\oval(\value{fstwidth},\value{tapeboxhight})}
       \put(\value{curx},\value{cury})
              {\makebox(0,0){$FST_1 \circ FST_2 \circ \cdots \circ FST_n$}}%
    \end{picture}%
   }

\newcommand{\paralleltransducers}%
   {\paralleltransducersone%
    ~$\Longrightarrow$~%
    \paralleltransducerstwo%
   }

\newcommand{\paralleltransducersone}%
   {
    \setcounter{picwidth}{10*\value{tapeboxhight}}%
    \setcounter{pichight}{9*\value{tapeboxhight}}%
    \setcounter{lift}{\value{pichight}/-2}%
    \rule[\the\value{lift}pt]{0 pt}{\the\value{pichight}pt}%
    \begin{picture}(\the\value{picwidth},0)(0,-\the\value{lift})%
       \put(0,0){\framebox(\value{picwidth},\value{tapeboxhight})
             {Surface String}}%
       \setcounter{cury}{8*\value{tapeboxhight}}
       \put(0,\value{cury}){\framebox(\value{picwidth},\value{tapeboxhight})
             {Lexical String}}%
       \setcounter{curx}{\value{picwidth}/2}
       \setcounter{cury}{\value{tapeboxhight}}
       \setcounter{delta}{6*\value{tapeboxhight}}
       \multiput(\value{curx},\value{cury})(0,\value{delta}){2}
          {\line(0,1){\value{tapeboxhight}}}
       \setcounter{curx}{\value{tapeboxhight}}
       \setcounter{cury}{2*\value{tapeboxhight}}
       \setcounter{delta}{5*\value{tapeboxhight}}
       \setcounter{temp}{8*\value{tapeboxhight}}
       \multiput(\value{curx},\value{cury})(0,\value{delta}){2}
          {\line(1,0){\value{temp}}}
       \setcounter{delta}{3*\value{tapeboxhight}}
       \setcounter{temp}{2*\value{tapeboxhight}}
       \multiput(\value{curx},\value{cury})(\value{delta},0){2}
          {\line(0,1){\value{temp}}}
       \setcounter{cury}{5*\value{tapeboxhight}}
       \multiput(\value{curx},\value{cury})(\value{delta},0){2}
          {\line(0,1){\value{temp}}}
       \setcounter{cury}{4*\value{tapeboxhight}+\value{tapeboxhight}/2}
       \setcounter{fstwidth}{2*\value{tapeboxhight}}
       \multiput(\value{curx},\value{cury})(\value{delta},0){2}
          {\oval(\value{fstwidth},\value{tapeboxhight})}
       \put(\value{curx},\value{cury}){\makebox(0,0){\em FST$_1$}}%
       \setcounter{curx}{4*\value{tapeboxhight}}
       \put(\value{curx},\value{cury}){\makebox(0,0){\em FST$_2$}}%
       \setcounter{curx}{9*\value{tapeboxhight}}
       \setcounter{cury}{2*\value{tapeboxhight}}
       \setcounter{delta}{3*\value{tapeboxhight}}
       \multiput(\value{curx},\value{cury})(0,\value{delta}){2}
          {\line(0,1){\value{temp}}}
       \setcounter{cury}{4*\value{tapeboxhight}+\value{tapeboxhight}/2}
       \put(\value{curx},\value{cury})
          {\oval(\value{fstwidth},\value{tapeboxhight})}
       \put(\value{curx},\value{cury}){\makebox(0,0){\em FST$_n$}}%
       \setcounter{curx}{6*\value{tapeboxhight}+\value{tapeboxhight}/2}
       \put(\value{curx},\value{cury}){\makebox(0,0){$\cdots$}}%
    \end{picture}%
   }
\newcommand{\paralleltransducerstwo}%
   {\begin{picture}(\the\value{picwidth},0)(0,-\the\value{lift})%
       \put(0,0){\framebox(\value{picwidth},\value{tapeboxhight})
             {Surface String}}%
       \setcounter{cury}{8*\value{tapeboxhight}}
       \put(0,\value{cury}){\framebox(\value{picwidth},\value{tapeboxhight})
             {Lexical String}}%
       \setcounter{curx}{\value{picwidth}/2}
       \setcounter{cury}{\value{tapeboxhight}}
       \setcounter{temp}{3*\value{tapeboxhight}}
       \setcounter{delta}{4*\value{tapeboxhight}}
       \multiput(\value{curx},\value{cury})(0,\value{delta}){2}
          {\line(0,1){\value{temp}}}
       \setcounter{cury}{4*\value{tapeboxhight}+\value{tapeboxhight}/2}
       \setcounter{fstwidth}{8*\value{tapeboxhight}}
       \put(\value{curx},\value{cury})
              {\oval(\value{fstwidth},\value{tapeboxhight})}
       \put(\value{curx},\value{cury})
              {\makebox(0,0){$FST_1 \cap FST_2 \cap \cdots \cap FST_n$}}%
    \end{picture}%
   }

\newcommand{\uniontransducers}%
   {
    \setcounter{picwidth}{10*\value{tapeboxhight}}%
    \setcounter{pichight}{9*\value{tapeboxhight}}%
    \setcounter{lift}{\value{pichight}/-2}%
    \rule[\the\value{lift}pt]{0 pt}{\the\value{pichight}pt}%
    \begin{picture}(\the\value{picwidth},0)(0,-\the\value{lift})%
       \put(0,0){\framebox(\value{picwidth},\value{tapeboxhight})
             {Surface String}}%
       \setcounter{curx}{2*\value{tapeboxhight}+\value{tapeboxhight}/2}
       \setcounter{delta}{2*\value{tapeboxhight}+\value{tapeboxhight}/2}
       \multiput(\value{curx},0)(\value{delta},0){3}
          {\dashbox{.75}(0,\value{tapeboxhight}){}}
       \setcounter{cury}{8*\value{tapeboxhight}}
       \put(0,\value{cury}){\framebox(\value{picwidth},\value{tapeboxhight})
             {Lexical String}}%
       \setcounter{cury}{8*\value{tapeboxhight}}
       \multiput(\value{curx},\value{cury})(\value{delta},0){3}
          {\dashbox{.75}(0,\value{tapeboxhight}){}}
       \setcounter{curx}{\value{tapeboxhight}}
       \setcounter{cury}{\value{tapeboxhight}}
       \setcounter{delta}{3*\value{tapeboxhight}}
       \multiput(\value{curx},\value{cury})(\value{delta},0){2}
          {\line(0,1){\value{delta}}}
       \setcounter{cury}{5*\value{tapeboxhight}}
       \multiput(\value{curx},\value{cury})(\value{delta},0){2}
          {\line(0,1){\value{delta}}}
       \setcounter{cury}{4*\value{tapeboxhight}+\value{tapeboxhight}/2}
       \setcounter{fstwidth}{2*\value{tapeboxhight}}
       \multiput(\value{curx},\value{cury})(\value{delta},0){2}
          {\oval(\value{fstwidth},\value{tapeboxhight})}
       \put(\value{curx},\value{cury}){\makebox(0,0){\em FST$_1$}}%
       \setcounter{curx}{4*\value{tapeboxhight}}
       \put(\value{curx},\value{cury}){\makebox(0,0){\em FST$_2$}}%
       \setcounter{curx}{9*\value{tapeboxhight}}
       \setcounter{cury}{\value{tapeboxhight}}
       \setcounter{delta}{4*\value{tapeboxhight}}
       \multiput(\value{curx},\value{cury})(0,\value{delta}){2}
          {\line(0,1){\value{temp}}}
       \setcounter{cury}{4*\value{tapeboxhight}+\value{tapeboxhight}/2}
       \put(\value{curx},\value{cury})
          {\oval(\value{fstwidth},\value{tapeboxhight})}
       \put(\value{curx},\value{cury}){\makebox(0,0){\em FST$_n$}}%
       \setcounter{curx}{6*\value{tapeboxhight}+\value{tapeboxhight}/2}
       \put(\value{curx},\value{cury}){\makebox(0,0){$\cdots$}}%
    \end{picture}%
    ~$\Longrightarrow$~%
    \begin{picture}(\the\value{picwidth},0)(0,-\the\value{lift})%
       \put(0,0){\framebox(\value{picwidth},\value{tapeboxhight})
             {Surface String}}%
       \setcounter{curx}{2*\value{tapeboxhight}+\value{tapeboxhight}/2}
       \setcounter{delta}{2*\value{tapeboxhight}+\value{tapeboxhight}/2}
       \multiput(\value{curx},0)(\value{delta},0){3}
          {\dashbox{.75}(0,\value{tapeboxhight}){}}
       \setcounter{cury}{8*\value{tapeboxhight}}
       \put(0,\value{cury}){\framebox(\value{picwidth},\value{tapeboxhight})
             {Lexical String}}%
       \setcounter{cury}{8*\value{tapeboxhight}}
       \multiput(\value{curx},\value{cury})(\value{delta},0){3}
          {\dashbox{.75}(0,\value{tapeboxhight}){}}
       \setcounter{curx}{\value{picwidth}/2}
       \setcounter{cury}{4*\value{tapeboxhight}+\value{tapeboxhight}/2}
       \setcounter{fstwidth}{8*\value{tapeboxhight}}
       \put(\value{curx},\value{cury})
              {\oval(\value{fstwidth},\value{tapeboxhight})}
       \put(\value{curx},\value{cury})
              {\makebox(0,0){$FST_1 \cup FST_2 \cup \cdots \cup FST_n$}}%
       \setcounter{cury}{5*\value{tapeboxhight}}
       \setcounter{temp}{4*\value{tapeboxhight}}
       \put(\value{curx},\value{cury}){\line(-4,3){\value{temp}}}
       \put(\value{curx},\value{cury}){\line(4,3){\value{temp}}}
       \put(\value{curx},\value{cury}){\line(-1,3){\value{tapeboxhight}}}
       \put(\value{curx},\value{cury}){\line(1,3){\value{tapeboxhight}}}
       \setcounter{cury}{4*\value{tapeboxhight}}
       \setcounter{temp}{4*\value{tapeboxhight}}
       \put(\value{curx},\value{cury}){\line(-4,-3){\value{temp}}}
       \put(\value{curx},\value{cury}){\line(4,-3){\value{temp}}}
       \put(\value{curx},\value{cury}){\line(-1,-3){\value{tapeboxhight}}}
       \put(\value{curx},\value{cury}){\line(1,-3){\value{tapeboxhight}}}
    \end{picture}%
   }

\newcommand{\katajakoskenniemi}%
   {
    \setcounter{picwidth}{10*\value{tapeboxhight}}%
    \setcounter{pichight}{9*\value{tapeboxhight}}%
    \setcounter{lift}{\value{pichight}/-2}%
    \rule[\the\value{lift}pt]{0 pt}{\the\value{pichight}pt}%
    \begin{picture}(\the\value{picwidth},0)(0,-\the\value{lift})%
       \put(0,0){\makebox(\value{picwidth},\value{tapeboxhight})
             {Surface Representation}}%
       \setcounter{cury}{4*\value{tapeboxhight}}
       \setcounter{temp}{8*\value{tapeboxhight}}
       \put(\value{tapeboxhight},\value{cury})
             {\makebox(\value{temp},\value{tapeboxhight})
             {Lexical Representation}}%
       \setcounter{cury}{8*\value{tapeboxhight}}
       \put(0,\value{cury}){\makebox(\value{picwidth},\value{tapeboxhight})
             {Lexical Entries (Morphemes)}}%
       \setcounter{curx}{\value{picwidth}/2}
       \setcounter{cury}{\value{tapeboxhight}}
       \setcounter{delta}{2*\value{tapeboxhight}}
       \multiput(\value{curx},\value{cury})(0,\value{delta}){4}
          {\line(0,1){\value{tapeboxhight}}}
       \setcounter{cury}{2*\value{tapeboxhight}+\value{tapeboxhight}/2}
       \setcounter{delta}{4*\value{tapeboxhight}}
       \setcounter{fstwidth}{10*\value{tapeboxhight}}
       \multiput(\value{curx},\value{cury})(0,\value{delta}){2}
          {\oval(\value{fstwidth},\value{tapeboxhight})}
       \put(\value{curx},\value{cury}){\makebox(0,0){\sc Two-Level Rules}}%
       \setcounter{cury}{6*\value{tapeboxhight}+\value{tapeboxhight}/2}
       \put(\value{curx},\value{cury}){\makebox(0,0){\sc Lexicon Component}}%
    \end{picture}%
   }

\newcommand{\environbar}{\underline{\hspace*{1.5em}}\ }

\newcommand{\phonrule}[4]%
   {#1 {}$\rightarrow${} #2 / #3 \environbar #4}

\newcommand{\tlr}[7]%
   {\begin{tabular}{cccccc}%
      {}#5&--&#6&--&#7&#4 \\
      {}#1&--&#2&--&#3&
   \end{tabular}%
   \vspace{.1in}}

\newcommand{\tlrf}[8]%
   {\begin{tabular}{cccccc}%
      {}#5&--&#6&--&#7&#4\\
      {}#1&--&#2&--&#3\\
      \multicolumn{6}{l}{{\sf Features:} {\tt #8}}
   \end{tabular}
   \vspace{.1in}}

\newcommand{\tlrc}[8]%
   {\begin{tabular}{cccccc}%
      {}#5&--&#6&--&#7&#4 \\
      {}#1&--&#2&--&#3& \\
      \multicolumn{6}{l}{{\sf where} #8}%
   \end{tabular}%
   \vspace{.1in}}

\newcommand{\tlrt}[8]%
   {\tlrc#1#2#3#4#5#6#7#8}

\newcommand{\tlrule}[9]
   {\begin{tabbing}%
       tl\_rule({\tt #1},
                  \= {\tt #2}, \= {\tt #3}, \= {\tt #4}, {\tt #5},\\%
                  \> {\tt #6}, \> {\tt #7}, \> {\tt #8},\\%
                  \> \restoftlrule#9END%
    \end{tabbing}%
   }

\newcommand{\tlruleintab}[9]
   {tl\_rule({\tt #1},
                  \= {\tt #2}, \= {\tt #3}, \= {\tt #4}, {\tt #5},\\%
             \>   \> {\tt #6}, \> {\tt #7}, \> {\tt #8},\\%
             \>   \> \restoftlrule#9END%
   }

\newcommand{\tlruleintablong}[9]
   {tl\_rule({\tt #1},
                  \= {\tt #2}, {\tt #3}, {\tt #4}, {\tt #5},
                     {\tt #6}, {\tt #7}, {\tt #8},\\%
              \>  \> \restoftlrule#9END%
   }

\def\restoftlrule#1|#2END%
  {{\tt #1}, {\tt #2}).}

 
\def\diatop[#1|#2]{{\setbox1=\hbox{{#1{}}}\setbox2=\hbox{{#2{}}}%
                    \dimen0=\ifdim\wd1>\wd2\wd1\else\wd2\fi%
                    \dimen1=\ht2\advance\dimen1by-1ex%
                    \setbox1=\hbox to1\dimen0{\hss#1\hss}%
                    \rlap{\raise1\dimen1\box1}%
                    \hbox to1\dimen0{\hss#2\hss}}}%
 
 
 
\font\ipatwelverm=wsuipa12
\def\ipa{\ipatwelverm}
 
\def\inva{\edef\next{\the\font}\ipa\char'000\next}%
\def\scripta{\edef\next{\the\font}\ipa\char'001\next}%
\def\nialpha{\edef\next{\the\font}\ipa\char'002\next}%
\def\invscripta{\edef\next{\the\font}\ipa\char'003\next}%
\def\invv{\edef\next{\the\font}\ipa\char'004\next}%
 
\def\crossb{\edef\next{\the\font}\ipa\char'005\next}%
\def\barb{\edef\next{\the\font}\ipa\char'006\next}%
\def\slashb{\edef\next{\the\font}\ipa\char'007\next}%
\def\hookb{\edef\next{\the\font}\ipa\char'010\next}%
\def\nibeta{\edef\next{\the\font}\ipa\char'011\next}%
 
\def\slashc{\edef\next{\the\font}\ipa\char'012\next}%
\def\curlyc{\edef\next{\the\font}\ipa\char'013\next}%
\def\clickc{\edef\next{\the\font}\ipa\char'014\next}%
 
\def\crossd{\edef\next{\the\font}\ipa\char'015\next}%
\def\bard{\edef\next{\the\font}\ipa\char'016\next}%
\def\slashd{\edef\next{\the\font}\ipa\char'017\next}%
\def\hookd{\edef\next{\the\font}\ipa\char'020\next}%
\def\taild{\edef\next{\the\font}\ipa\char'021\next}%
\def\dz{\edef\next{\the\font}\ipa\char'022\next}%
\def\eth{\edef\next{\the\font}\ipa\char'023\next}%
\def\scd{\edef\next{\the\font}\ipa\char'024\next}%
 
\def\schwa{\edef\next{\the\font}\ipa\char'025\next}%
\def\er{\edef\next{\the\font}\ipa\char'026\next}%
\def\reve{\edef\next{\the\font}\ipa\char'027\next}%
\def\niepsilon{\edef\next{\the\font}\ipa\char'030\next}%
\def\revepsilon{\edef\next{\the\font}\ipa\char'031\next}%
\def\hookrevepsilon{\edef\next{\the\font}\ipa\char'032\next}%
\def\closedrevepsilon{\edef\next{\the\font}\ipa\char'033\next}%
 
\def\scriptg{\edef\next{\the\font}\ipa\char'034\next}%
\def\hookg{\edef\next{\the\font}\ipa\char'035\next}%
\def\scg{\edef\next{\the\font}\ipa\char'036\next}%
\def\nigamma{\edef\next{\the\font}\ipa\char'037\next}
\def\ipagamma{\edef\next{\the\font}\ipa\char'040\next}%
\def\babygamma{\edef\next{\the\font}\ipa\char'041\next}%
 
\def\hv{\edef\next{\the\font}\ipa\char'042\next}%
\def\crossh{\edef\next{\the\font}\ipa\char'043\next}%
\def\hookh{\edef\next{\the\font}\ipa\char'044\next}%
\def\hookheng{\edef\next{\the\font}\ipa\char'045\next}%
\def\invh{\edef\next{\the\font}\ipa\char'046\next}%
 
\def\bari{\edef\next{\the\font}\ipa\char'047\next}%
\def\dlbari{\edef\next{\the\font}\ipa\char'050\next}
\def\niiota{\edef\next{\the\font}\ipa\char'051\next}%
\def\sci{\edef\next{\the\font}\ipa\char'052\next}%
\def\barsci{\edef\next{\the\font}\ipa\char'053\next}
 
\def\invf{\edef\next{\the\font}\ipa\char'054\next}%
 
\def\tildel{\edef\next{\the\font}\ipa\char'055\next}%
\def\barl{\edef\next{\the\font}\ipa\char'056\next}%
\def\latfric{\edef\next{\the\font}\ipa\char'057\next}%
\def\taill{\edef\next{\the\font}\ipa\char'060\next}%
\def\lz{\edef\next{\the\font}\ipa\char'061\next}%
\def\nilambda{\edef\next{\the\font}\ipa\char'062\next}%
\def\crossnilambda{\edef\next{\the\font}\ipa\char'063\next}%
 
\def\labdentalnas{\edef\next{\the\font}\ipa\char'064\next}%
\def\invm{\edef\next{\the\font}\ipa\char'065\next}%
\def\legm{\edef\next{\the\font}\ipa\char'066\next}%
 
\def\nj{\edef\next{\the\font}\ipa\char'067\next}%
\def\eng{\edef\next{\the\font}\ipa\char'070\next}%
\def\tailn{\edef\next{\the\font}\ipa\char'071\next}%
\def\scn{\edef\next{\the\font}\ipa\char'072\next}%
 
\def\clickb{\edef\next{\the\font}\ipa\char'073\next}%
\def\baro{\edef\next{\the\font}\ipa\char'074\next}%
\def\openo{\edef\next{\the\font}\ipa\char'075\next}%
\def\niomega{\edef\next{\the\font}\ipa\char'076\next}%
\def\closedniomega{\edef\next{\the\font}\ipa\char'077\next}%
\def\oo{\edef\next{\the\font}\ipa\char'100\next}%
 
\def\barp{\edef\next{\the\font}\ipa\char'101\next}%
\def\thorn{\edef\next{\the\font}\ipa\char'102\next}%
\def\niphi{\edef\next{\the\font}\ipa\char'103\next}%
 
\def\flapr{\edef\next{\the\font}\ipa\char'104\next}%
\def\legr{\edef\next{\the\font}\ipa\char'105\next}%
\def\tailr{\edef\next{\the\font}\ipa\char'106\next}%
\def\invr{\edef\next{\the\font}\ipa\char'107\next}%
\def\tailinvr{\edef\next{\the\font}\ipa\char'110\next}%
\def\invlegr{\edef\next{\the\font}\ipa\char'111\next}%
\def\scr{\edef\next{\the\font}\ipa\char'112\next}%
\def\invscr{\edef\next{\the\font}\ipa\char'113\next}%
 
\def\tails{\edef\next{\the\font}\ipa\char'114\next}%
\def\esh{\edef\next{\the\font}\ipa\char'115\next}%
\def\curlyesh{\edef\next{\the\font}\ipa\char'116\next}%
\def\nisigma{\edef\next{\the\font}\ipa\char'117\next}%
 
\def\tailt{\edef\next{\the\font}\ipa\char'120\next}%
\def\tesh{\edef\next{\the\font}\ipa\char'121\next}%
\def\clickt{\edef\next{\the\font}\ipa\char'122\next}%
\def\nitheta{\edef\next{\the\font}\ipa\char'123\next}%
 
\def\baru{\edef\next{\the\font}\ipa\char'124\next}%
\def\slashu{\edef\next{\the\font}\ipa\char'125\next}%
\def\niupsilon{\edef\next{\the\font}\ipa\char'126\next}%
\def\scu{\edef\next{\the\font}\ipa\char'127\next}%
\def\barscu{\edef\next{\the\font}\ipa\char'130\next}%
 
\def\scriptv{\edef\next{\the\font}\ipa\char'131\next}%
 
\def\invw{\edef\next{\the\font}\ipa\char'132\next}%
 
\def\nichi{\edef\next{\the\font}\ipa\char'133\next}%
 
\def\invy{\edef\next{\the\font}\ipa\char'134\next}%
\def\scy{\edef\next{\the\font}\ipa\char'135\next}%
 
\def\curlyz{\edef\next{\the\font}\ipa\char'136\next}%
\def\tailz{\edef\next{\the\font}\ipa\char'137\next}%
\def\yogh{\edef\next{\the\font}\ipa\char'140\next}%
\def\curlyyogh{\edef\next{\the\font}\ipa\char'141\next}%
 
\def\glotstop{\edef\next{\the\font}\ipa\char'142\next}%
\def\revglotstop{\edef\next{\the\font}\ipa\char'143\next}%
\def\invglotstop{\edef\next{\the\font}\ipa\char'144\next}%
\def\ejective{\edef\next{\the\font}\ipa\char'145\next}%
\def\reveject{\edef\next{\the\font}\ipa\char'146\next}%
 
 
\def\dental#1{\oalign{#1\crcr
          \hidewidth{\ipa\char'147}\hidewidth}}
 
\def\upt{\edef\next{\the\font}\ipa\char'154\next}
\def\downt{\edef\next{\the\font}\ipa\char'155\next}%
\def\leftt{\edef\next{\the\font}\ipa\char'156\next}%
\def\rightt{\edef\next{\the\font}\ipa\char'157\next}%
 
\def\upp{\edef\next{\the\font}\ipa\char'164\next}
\def\downp{\edef\next{\the\font}\ipa\char'165\next}%
\def\leftp{\edef\next{\the\font}\ipa\char'166\next}%
\def\rightp{\edef\next{\the\font}\ipa\char'167\next}%
 
\def\stress{\edef\next{\the\font}\ipa\char'150\next}
\def\secstress{\edef\next{\the\font}\ipa\char'151\next}
 
\def\syllabic{\edef\next{\the\font}\ipa\char'152\next}
 
\def\corner{\edef\next{\the\font}\ipa\char'153\next}%
 
\def\halflength{\edef\next{\the\font}\ipa\char'160\next}
\def\length{\edef\next{\the\font}\ipa\char'161\next}
 
\def\underdots{\edef\next{\the\font}\ipa\char'162\next}%
 
\def\ain{\edef\next{\the\font}\ipa\char'163\next}
 
\def\overring{\edef\next{\the\font}\ipa\char'170\next}%
\def\underring{\edef\next{\the\font}\ipa\char'171\next}%
 
\def\open{\edef\next{\the\font}\ipa\char'172\next}%
 
\def\midtilde{\edef\next{\the\font}\ipa\char'173\next}%
\def\undertilde{\edef\next{\the\font}\ipa\char'174\next}%
 
\def\underwedge{\edef\next{\the\font}\ipa\char'175\next}%
 
\def\polishhook{\edef\next{\the\font}\ipa\char'176\next}%
 
\def\underarch#1{\oalign{#1\crcr
          \hidewidth{\ipa\char'177}\hidewidth}}
 

\font\ipatenrm=wsuipa10
\def\ipa{\ipatenrm}

\newcommand{\A}{\ejective}  
\newcommand{\h}{\crossh}    
\newcommand{\CC}{\reveject} 
\newcommand{\sh}{\v{s}}	    
\newcommand{\J}{\^{\j}}     
\newcommand{\TT}{\d{t}}   
\newcommand{\Ss}{\d{s}}   
\newcommand{\OO}{\^{o}}      

\newcommand{\br}{\b{b}}   
\newcommand{\gr}{\b{g}}
\newcommand{\dr}{\b{d}}
\newcommand{\kr}{\b{k}}
\newcommand{\pr}{\b{p}}
\newcommand{\tr}{\b{t}}
\newcommand{\e}{\schwa}


\newcommand{\Ru}{{\em Rukk\={a}\kr\^{a}}}  
\newcommand{\Qu}{{\em Qu\sh\sh\={a}y\^{a}}}  
\newcommand{\bgdkpt}{{\em b\gr\={a}\dr k\pr\={a}\tr}}

\def\PsfigVersion{1.9}
\ifx\undefined\psfig\else\endinput\fi

%

\let\LaTeXAtSign=\@
\let\@=\relax
\edef\psfigRestoreAt{\catcode`\@=\number\catcode`@\relax}
\catcode`\@=11\relax
\newwrite\@unused
\def\ps@typeout#1{{\let\protect\string\immediate\write\@unused{#1}}}
\ps@typeout{psfig/tex \PsfigVersion}


\def\figurepath{./}
\def\psfigurepath#1{\edef\figurepath{#1}}

%
%
\def\@nnil{\@nil}
\def\@empty{}
\def\@psdonoop#1\@@#2#3{}
\def\@psdo#1:=#2\do#3{\edef\@psdotmp{#2}\ifx\@psdotmp\@empty \else
    \expandafter\@psdoloop#2,\@nil,\@nil\@@#1{#3}\fi}
\def\@psdoloop#1,#2,#3\@@#4#5{\def#4{#1}\ifx #4\@nnil \else
       #5\def#4{#2}\ifx #4\@nnil \else#5\@ipsdoloop #3\@@#4{#5}\fi\fi}
\def\@ipsdoloop#1,#2\@@#3#4{\def#3{#1}\ifx #3\@nnil 
       \let\@nextwhile=\@psdonoop \else
      #4\relax\let\@nextwhile=\@ipsdoloop\fi\@nextwhile#2\@@#3{#4}}
\def\@tpsdo#1:=#2\do#3{\xdef\@psdotmp{#2}\ifx\@psdotmp\@empty \else
    \@tpsdoloop#2\@nil\@nil\@@#1{#3}\fi}
\def\@tpsdoloop#1#2\@@#3#4{\def#3{#1}\ifx #3\@nnil 
       \let\@nextwhile=\@psdonoop \else
      #4\relax\let\@nextwhile=\@tpsdoloop\fi\@nextwhile#2\@@#3{#4}}
%
\ifx\undefined\fbox
\newdimen\fboxrule
\newdimen\fboxsep
\newdimen\ps@tempdima
\newbox\ps@tempboxa
\fboxsep = 3pt
\fboxrule = .4pt
\long\def\fbox#1{\leavevmode\setbox\ps@tempboxa\hbox{#1}\ps@tempdima\fboxrule
    \advance\ps@tempdima \fboxsep \advance\ps@tempdima \dp\ps@tempboxa
   \hbox{\lower \ps@tempdima\hbox
  {\vbox{\hrule height \fboxrule
          \hbox{\vrule width \fboxrule \hskip\fboxsep
          \vbox{\vskip\fboxsep \box\ps@tempboxa\vskip\fboxsep}\hskip 
                 \fboxsep\vrule width \fboxrule}
                 \hrule height \fboxrule}}}}
\fi
%
%
\newread\ps@stream
\newif\ifnot@eof       
\newif\if@noisy        
\newif\if@atend        
\newif\if@psfile       
%
%
{\catcode`\%=12\global\gdef\epsf@start{
\def\epsf@PS{PS}
\def\epsf@getbb#1{%
%
%
\openin\ps@stream=#1
\ifeof\ps@stream\ps@typeout{Error, File #1 not found}\else
%
%
   {\not@eoftrue \chardef\other=12
    \def\do##1{\catcode`##1=\other}\dospecials \catcode`\ =10
    \loop
       \if@psfile
	  \read\ps@stream to \epsf@fileline
       \else{
	  \obeyspaces
          \read\ps@stream to \epsf@tmp\global\let\epsf@fileline\epsf@tmp}
       \fi
       \ifeof\ps@stream\not@eoffalse\else
%
%
       \if@psfile\else
       \expandafter\epsf@test\epsf@fileline:. \\%
       \fi
%
%
          \expandafter\epsf@aux\epsf@fileline:. \\%
       \fi
   \ifnot@eof\repeat
   }\closein\ps@stream\fi}%
%
%
\long\def\epsf@test#1#2#3:#4\\{\def\epsf@testit{#1#2}
			\ifx\epsf@testit\epsf@start\else
\ps@typeout{Warning! File does not start with `\epsf@start'.  It may not be a PostScript file.}
			\fi
			\@psfiletrue} 
%
%
{\catcode`\%=12\global\let\epsf@percent=
%
%
%
\long\def\epsf@aux#1#2:#3\\{\ifx#1\epsf@percent
   \def\epsf@testit{#2}\ifx\epsf@testit\epsf@bblit
	\@atendfalse
        \epsf@atend #3 . \\%
	\if@atend	
	   \if@verbose{
		\ps@typeout{psfig: found `(atend)'; continuing search}
	   }\fi
        \else
        \epsf@grab #3 . . . \\%
        \not@eoffalse
        \global\no@bbfalse
        \fi
   \fi\fi}%
%
%
\def\epsf@grab #1 #2 #3 #4 #5\\{%
   \global\def\epsf@llx{#1}\ifx\epsf@llx\empty
      \epsf@grab #2 #3 #4 #5 .\\\else
   \global\def\epsf@lly{#2}%
   \global\def\epsf@urx{#3}\global\def\epsf@ury{#4}\fi}%
%
%
\def\epsf@atendlit{(atend)} 
\def\epsf@atend #1 #2 #3\\{%
   \def\epsf@tmp{#1}\ifx\epsf@tmp\empty
      \epsf@atend #2 #3 .\\\else
   \ifx\epsf@tmp\epsf@atendlit\@atendtrue\fi\fi}


\chardef\psletter = 11 
\chardef\other = 12

\newif \ifdebug 
\newif\ifc@mpute 
\c@mputetrue 

\let\then = \relax
\def\r@dian{pt }
\let\r@dians = \r@dian
\let\dimensionless@nit = \r@dian
\let\dimensionless@nits = \dimensionless@nit
\def\internal@nit{sp }
\let\internal@nits = \internal@nit
\newif\ifstillc@nverging
\def \Mess@ge #1{\ifdebug \then \message {#1} \fi}

{ 
	\catcode `\@ = \psletter
	\gdef \nodimen {\expandafter \n@dimen \the \dimen}
	\gdef \term #1 #2 #3%
	       {\edef \t@ {\the #1}
		\edef \t@@ {\expandafter \n@dimen \the #2\r@dian}%
		\t@rm {\t@} {\t@@} {#3}%
	       }
	\gdef \t@rm #1 #2 #3%
	       {{%
		\count 0 = 0
		\dimen 0 = 1 \dimensionless@nit
		\dimen 2 = #2\relax
		\Mess@ge {Calculating term #1 of \nodimen 2}%
		\loop
		\ifnum	\count 0 < #1
		\then	\advance \count 0 by 1
			\Mess@ge {Iteration \the \count 0 \space}%
			\Multiply \dimen 0 by {\dimen 2}%
			\Mess@ge {After multiplication, term = \nodimen 0}%
			\Divide \dimen 0 by {\count 0}%
			\Mess@ge {After division, term = \nodimen 0}%
		\repeat
		\Mess@ge {Final value for term #1 of 
				\nodimen 2 \space is \nodimen 0}%
		\xdef \Term {#3 = \nodimen 0 \r@dians}%
		\aftergroup \Term
	       }}
	\catcode `\p = \other
	\catcode `\t = \other
	\gdef \n@dimen #1pt{#1} 
}

\def \Divide #1by #2{\divide #1 by #2} 

\def \Multiply #1by #2
       {{
	\count 0 = #1\relax
	\count 2 = #2\relax
	\count 4 = 65536
	\Mess@ge {Before scaling, count 0 = \the \count 0 \space and
			count 2 = \the \count 2}%
	\ifnum	\count 0 > 32767 
	\then	\divide \count 0 by 4
		\divide \count 4 by 4
	\else	\ifnum	\count 0 < -32767
		\then	\divide \count 0 by 4
			\divide \count 4 by 4
		\else
		\fi
	\fi
	\ifnum	\count 2 > 32767 
	\then	\divide \count 2 by 4
		\divide \count 4 by 4
	\else	\ifnum	\count 2 < -32767
		\then	\divide \count 2 by 4
			\divide \count 4 by 4
		\else
		\fi
	\fi
	\multiply \count 0 by \count 2
	\divide \count 0 by \count 4
	\xdef \product {#1 = \the \count 0 \internal@nits}%
	\aftergroup \product
       }}

\def\r@duce{\ifdim\dimen0 > 90\r@dian \then   
		\multiply\dimen0 by -1
		\advance\dimen0 by 180\r@dian
		\r@duce
	    \else \ifdim\dimen0 < -90\r@dian \then  
		\advance\dimen0 by 360\r@dian
		\r@duce
		\fi
	    \fi}

\def\Sine#1%
       {{%
	\dimen 0 = #1 \r@dian
	\r@duce
	\ifdim\dimen0 = -90\r@dian \then
	   \dimen4 = -1\r@dian
	   \c@mputefalse
	\fi
	\ifdim\dimen0 = 90\r@dian \then
	   \dimen4 = 1\r@dian
	   \c@mputefalse
	\fi
	\ifdim\dimen0 = 0\r@dian \then
	   \dimen4 = 0\r@dian
	   \c@mputefalse
	\fi
	\ifc@mpute \then
		\divide\dimen0 by 180
		\dimen0=3.141592654\dimen0
		\dimen 2 = 3.1415926535897963\r@dian 
		\divide\dimen 2 by 2 
		\Mess@ge {Sin: calculating Sin of \nodimen 0}%
		\count 0 = 1 
		\dimen 2 = 1 \r@dian 
		\dimen 4 = 0 \r@dian 
		\loop
			\ifnum	\dimen 2 = 0 
			\then	\stillc@nvergingfalse 
			\else	\stillc@nvergingtrue
			\fi
			\ifstillc@nverging 
			\then	\term {\count 0} {\dimen 0} {\dimen 2}%
				\advance \count 0 by 2
				\count 2 = \count 0
				\divide \count 2 by 2
				\ifodd	\count 2 
				\then	\advance \dimen 4 by \dimen 2
				\else	\advance \dimen 4 by -\dimen 2
				\fi
		\repeat
	\fi		
			\xdef \sine {\nodimen 4}%
       }}

\def\Cosine#1{\ifx\sine\UnDefined\edef\Savesine{\relax}\else
		             \edef\Savesine{\sine}\fi
	{\dimen0=#1\r@dian\advance\dimen0 by 90\r@dian
	 \Sine{\nodimen 0}
	 \xdef\cosine{\sine}
	 \xdef\sine{\Savesine}}}	      

\def\psdraft{
	\def\@psdraft{0}
}
\def\psfull{
	\def\@psdraft{100}
}

\psfull

\newif\if@scalefirst
\def\psscalefirst{\@scalefirsttrue}
\def\psrotatefirst{\@scalefirstfalse}
\psrotatefirst

\newif\if@draftbox
\def\psnodraftbox{
	\@draftboxfalse
}
\def\psdraftbox{
	\@draftboxtrue
}
\@draftboxtrue

\newif\if@prologfile
\newif\if@postlogfile
\def\pssilent{
	\@noisyfalse
}
\def\psnoisy{
	\@noisytrue
}
\psnoisy
\newif\if@bbllx
\newif\if@bblly
\newif\if@bburx
\newif\if@bbury
\newif\if@height
\newif\if@width
\newif\if@rheight
\newif\if@rwidth
\newif\if@angle
\newif\if@clip
\newif\if@verbose
\def\@p@@sclip#1{\@cliptrue}

\newif\if@decmpr


\def\@p@@sfigure#1{\def\@p@sfile{null}\def\@p@sbbfile{null}
	        \openin1=#1.bb
		\ifeof1\closein1
	        	\openin1=\figurepath#1.bb
			\ifeof1\closein1
			        \openin1=#1
				\ifeof1\closein1%
				       \openin1=\figurepath#1
					\ifeof1
					   \ps@typeout{Error, File #1 not found}
						\if@bbllx\if@bblly
				   		\if@bburx\if@bbury
			      				\def\@p@sfile{#1}%
			      				\def\@p@sbbfile{#1}%
							\@decmprfalse
				  	   	\fi\fi\fi\fi
					\else\closein1
				    		\def\@p@sfile{\figurepath#1}%
				    		\def\@p@sbbfile{\figurepath#1}%
						\@decmprfalse
	                       		\fi%
			 	\else\closein1%
					\def\@p@sfile{#1}
					\def\@p@sbbfile{#1}
					\@decmprfalse
			 	\fi
			\else
				\def\@p@sfile{\figurepath#1}
				\def\@p@sbbfile{\figurepath#1.bb}
				\@decmprtrue
			\fi
		\else
			\def\@p@sfile{#1}
			\def\@p@sbbfile{#1.bb}
			\@decmprtrue
		\fi}

\def\@p@@sfile#1{\@p@@sfigure{#1}}

\def\@p@@sbbllx#1{
		\@bbllxtrue
		\dimen100=#1
		\edef\@p@sbbllx{\number\dimen100}
}
\def\@p@@sbblly#1{
		\@bbllytrue
		\dimen100=#1
		\edef\@p@sbblly{\number\dimen100}
}
\def\@p@@sbburx#1{
		\@bburxtrue
		\dimen100=#1
		\edef\@p@sbburx{\number\dimen100}
}
\def\@p@@sbbury#1{
		\@bburytrue
		\dimen100=#1
		\edef\@p@sbbury{\number\dimen100}
}
\def\@p@@sheight#1{
		\@heighttrue
		\dimen100=#1
   		\edef\@p@sheight{\number\dimen100}
}
\def\@p@@swidth#1{
		\@widthtrue
		\dimen100=#1
		\edef\@p@swidth{\number\dimen100}
}
\def\@p@@srheight#1{
		\@rheighttrue
		\dimen100=#1
		\edef\@p@srheight{\number\dimen100}
}
\def\@p@@srwidth#1{
		\@rwidthtrue
		\dimen100=#1
		\edef\@p@srwidth{\number\dimen100}
}
\def\@p@@sangle#1{
		\@angletrue
		\edef\@p@sangle{#1} 
}
\def\@p@@ssilent#1{ 
		\@verbosefalse
}
\def\@p@@sprolog#1{\@prologfiletrue\def\@prologfileval{#1}}
\def\@p@@spostlog#1{\@postlogfiletrue\def\@postlogfileval{#1}}
\def\@cs@name#1{\csname #1\endcsname}
\def\@setparms#1=#2,{\@cs@name{@p@@s#1}{#2}}
%
%
\def\ps@init@parms{
		\@bbllxfalse \@bbllyfalse
		\@bburxfalse \@bburyfalse
		\@heightfalse \@widthfalse
		\@rheightfalse \@rwidthfalse
		\def\@p@sbbllx{}\def\@p@sbblly{}
		\def\@p@sbburx{}\def\@p@sbbury{}
		\def\@p@sheight{}\def\@p@swidth{}
		\def\@p@srheight{}\def\@p@srwidth{}
		\def\@p@sangle{0}
		\def\@p@sfile{} \def\@p@sbbfile{}
		\def\@p@scost{10}
		\def\@sc{}
		\@prologfilefalse
		\@postlogfilefalse
		\@clipfalse
		\if@noisy
			\@verbosetrue
		\else
			\@verbosefalse
		\fi
}
%
%
\def\parse@ps@parms#1{
	 	\@psdo\@psfiga:=#1\do
		   {\expandafter\@setparms\@psfiga,}}
%
%
\newif\ifno@bb
\def\bb@missing{
	\if@verbose{
		\ps@typeout{psfig: searching \@p@sbbfile \space  for bounding box}
	}\fi
	\no@bbtrue
	\epsf@getbb{\@p@sbbfile}
        \ifno@bb \else \bb@cull\epsf@llx\epsf@lly\epsf@urx\epsf@ury\fi
}	
\def\bb@cull#1#2#3#4{
	\dimen100=#1 bp\edef\@p@sbbllx{\number\dimen100}
	\dimen100=#2 bp\edef\@p@sbblly{\number\dimen100}
	\dimen100=#3 bp\edef\@p@sbburx{\number\dimen100}
	\dimen100=#4 bp\edef\@p@sbbury{\number\dimen100}
	\no@bbfalse
}
\newdimen\p@intvaluex
\newdimen\p@intvaluey
\def\rotate@#1#2{{\dimen0=#1 sp\dimen1=#2 sp
		  \global\p@intvaluex=\cosine\dimen0
		  \dimen3=\sine\dimen1
		  \global\advance\p@intvaluex by -\dimen3
		  \global\p@intvaluey=\sine\dimen0
		  \dimen3=\cosine\dimen1
		  \global\advance\p@intvaluey by \dimen3
		  }}
\def\compute@bb{
		\no@bbfalse
		\if@bbllx \else \no@bbtrue \fi
		\if@bblly \else \no@bbtrue \fi
		\if@bburx \else \no@bbtrue \fi
		\if@bbury \else \no@bbtrue \fi
		\ifno@bb \bb@missing \fi
		\ifno@bb \ps@typeout{FATAL ERROR: no bb supplied or found}
			\no-bb-error
		\fi
		%
%
		\count203=\@p@sbburx
		\count204=\@p@sbbury
		\advance\count203 by -\@p@sbbllx
		\advance\count204 by -\@p@sbblly
		\edef\ps@bbw{\number\count203}
		\edef\ps@bbh{\number\count204}
		\if@angle 
			\Sine{\@p@sangle}\Cosine{\@p@sangle}
	        	{\dimen100=\maxdimen\xdef\r@p@sbbllx{\number\dimen100}
					    \xdef\r@p@sbblly{\number\dimen100}
			                    \xdef\r@p@sbburx{-\number\dimen100}
					    \xdef\r@p@sbbury{-\number\dimen100}}
%
                        \def\minmaxtest{
			   \ifnum\number\p@intvaluex<\r@p@sbbllx
			      \xdef\r@p@sbbllx{\number\p@intvaluex}\fi
			   \ifnum\number\p@intvaluex>\r@p@sbburx
			      \xdef\r@p@sbburx{\number\p@intvaluex}\fi
			   \ifnum\number\p@intvaluey<\r@p@sbblly
			      \xdef\r@p@sbblly{\number\p@intvaluey}\fi
			   \ifnum\number\p@intvaluey>\r@p@sbbury
			      \xdef\r@p@sbbury{\number\p@intvaluey}\fi
			   }
			\rotate@{\@p@sbbllx}{\@p@sbblly}
			\minmaxtest
			\rotate@{\@p@sbbllx}{\@p@sbbury}
			\minmaxtest
			\rotate@{\@p@sbburx}{\@p@sbblly}
			\minmaxtest
			\rotate@{\@p@sbburx}{\@p@sbbury}
			\minmaxtest
			\edef\@p@sbbllx{\r@p@sbbllx}\edef\@p@sbblly{\r@p@sbblly}
			\edef\@p@sbburx{\r@p@sbburx}\edef\@p@sbbury{\r@p@sbbury}
		\fi
		\count203=\@p@sbburx
		\count204=\@p@sbbury
		\advance\count203 by -\@p@sbbllx
		\advance\count204 by -\@p@sbblly
		\edef\@bbw{\number\count203}
		\edef\@bbh{\number\count204}
}
%
%
\def\in@hundreds#1#2#3{\count240=#2 \count241=#3
		     \count100=\count240	
		     \divide\count100 by \count241
		     \count101=\count100
		     \multiply\count101 by \count241
		     \advance\count240 by -\count101
		     \multiply\count240 by 10
		     \count101=\count240	
		     \divide\count101 by \count241
		     \count102=\count101
		     \multiply\count102 by \count241
		     \advance\count240 by -\count102
		     \multiply\count240 by 10
		     \count102=\count240	
		     \divide\count102 by \count241
		     \count200=#1\count205=0
		     \count201=\count200
			\multiply\count201 by \count100
		 	\advance\count205 by \count201
		     \count201=\count200
			\divide\count201 by 10
			\multiply\count201 by \count101
			\advance\count205 by \count201
		     \count201=\count200
			\divide\count201 by 100
			\multiply\count201 by \count102
			\advance\count205 by \count201
		     \edef\@result{\number\count205}
}
\def\compute@wfromh{
		\in@hundreds{\@p@sheight}{\@bbw}{\@bbh}
		\edef\@p@swidth{\@result}
}
\def\compute@hfromw{
	        \in@hundreds{\@p@swidth}{\@bbh}{\@bbw}
		\edef\@p@sheight{\@result}
}
\def\compute@handw{
		\if@height 
			\if@width
			\else
				\compute@wfromh
			\fi
		\else 
			\if@width
				\compute@hfromw
			\else
				\edef\@p@sheight{\@bbh}
				\edef\@p@swidth{\@bbw}
			\fi
		\fi
}
\def\compute@resv{
		\if@rheight \else \edef\@p@srheight{\@p@sheight} \fi
		\if@rwidth \else \edef\@p@srwidth{\@p@swidth} \fi
}
%
\def\compute@sizes{
	\compute@bb
	\if@scalefirst\if@angle
	\if@width
	   \in@hundreds{\@p@swidth}{\@bbw}{\ps@bbw}
	   \edef\@p@swidth{\@result}
	\fi
	\if@height
	   \in@hundreds{\@p@sheight}{\@bbh}{\ps@bbh}
	   \edef\@p@sheight{\@result}
	\fi
	\fi\fi
	\compute@handw
	\compute@resv}

%
%
\def\psfig#1{\vbox {
	%
	\ps@init@parms
	\parse@ps@parms{#1}
	\compute@sizes
	\ifnum\@p@scost<\@psdraft{
		\special{ps::[begin] 	\@p@swidth \space \@p@sheight \space
				\@p@sbbllx \space \@p@sbblly \space
				\@p@sbburx \space \@p@sbbury \space
				startTexFig \space }
		\if@angle
			\special {ps:: \@p@sangle \space rotate \space} 
		\fi
		\if@clip{
			\if@verbose{
				\ps@typeout{(clip)}
			}\fi
			\special{ps:: doclip \space }
		}\fi
		\if@prologfile
		    \special{ps: plotfile \@prologfileval \space } \fi
		\if@decmpr{
			\if@verbose{
				\ps@typeout{psfig: including \@p@sfile.Z \space }
			}\fi
			\special{ps: plotfile "`zcat \@p@sfile.Z" \space }
		}\else{
			\if@verbose{
				\ps@typeout{psfig: including \@p@sfile \space }
			}\fi
			\special{ps: plotfile \@p@sfile \space }
		}\fi
		\if@postlogfile
		    \special{ps: plotfile \@postlogfileval \space } \fi
		\special{ps::[end] endTexFig \space }
		\vbox to \@p@srheight sp{
			\hbox to \@p@srwidth sp{
				\hss
			}
		\vss
		}
	}\else{
		\if@draftbox{		
			\hbox{\frame{\vbox to \@p@srheight sp{
			\vss
			\hbox to \@p@srwidth sp{ \hss \@p@sfile \hss }
			\vss
			}}}
		}\else{
			\vbox to \@p@srheight sp{
			\vss
			\hbox to \@p@srwidth sp{\hss}
			\vss
			}
		}\fi

	}\fi
}}
\psfigRestoreAt
\let\@=\LaTeXAtSign

%

\newcounter{examplectr}
\newcounter{subexamplectr}
%
\newenvironment{ex}%
   {\addtocounter{examplectr}{1}%
     \setcounter{subexamplectr}{0}%
    \begin{list}{(\arabic{examplectr})}%
                {\setlength{\topsep}{0in}
	         \setlength{\labelsep}{0.075in}}
       \item
   }%
   {\end{list}%
    \medskip}
%
\newenvironment{subex}%
   {\addtocounter{subexamplectr}{1}
    \begin{list}{\alph{subexamplectr}.}%
                {\setlength{\topsep}{-\parskip}
	         \setlength{\labelsep}{0.075in}}
       \item \begin{minipage}[t]{5.5in}
   }%
   {\end{minipage}%
    \end{list}%
    \medskip}

%
\newcommand{\exnum}[2]{\addtocounter{examplectr}{#1}(\arabic{examplectr}{#2})\addtocounter{examplectr}{-#1}}

\newcommand{\Xbar}{$\overline{\mbox{X}}$}


\begin{abstract}
This paper presents a generalised two-level implementation
which can handle linear and non-linear morphological operations. An
algorithm for the interpretation of multi-tape two-level rules is
described. In addition, a number of issues which arise when developing
non-linear grammars are discussed with examples from Syriac.
\end{abstract}

\section{Introduction}

The introduction of two-level morphology \cite{Koskenniemi:83} and
subsequent developments has made implementing computational-morphology
models a feasible task. Yet, two-level formalisms fell short from
providing elegant means for the description of non-linear operations
such as infixation, circumfixation and root-and-pattern
morphology.\footnote{Although it is possible to express some classes
of non-linear rules using standard two-level formalisms by means of
{\em ad hoc} diacritics, e.g., infixation in
\cite[p.~156]{Antworth:90}, there are no means for expressing other
classes as root-and-pattern phenomena.} As a result, two-level
implementations --
e.g. \cite{Antworth:90,Karttunen:83,Karttunen:92,Ritchie:92book} --
have always been biased towards linear morphology.

The past decade has seen a number of proposals for handling non-linear
morphology;\footnote{\cite{Kay:87}, \cite{Kataja:88},
\cite{Beesley:89}, \cite{Lavie:90}, \cite{Beesley:90},
\cite{Beesley:91}, \cite{Kornai:91}, \cite{Wiebe:92},
\cite{Pulman:93}, \cite{Narayanan:93}, and \cite{Bird:94}. See
\cite{Kiraz:thesis} for a review.} however, none (apart from
Beesley's work) seem to have been implemented over large
descriptions, nor have they provided means by which the grammarian can
develop non-linear descriptions using higher level notation.

To test the validity of one's proposal or formalism, minimally a
medium-scale description is a desideratum. {\tt SemHe}\footnote{The
name {\tt SemHe} (Syriac {\em \d{s}em\d{h}\^{e}} `rays') is not an
acronym, but the title of a grammatical treatise written by the Syriac
polymath ({\em inter alia} mathematician and grammarian) Bar
`E\b{b}r\={o}y\^{o} (1225-1286), viz.~{\em k\b{t}\={o}\b{b}\^{o}
\b{d}\d{s}em\d{h}\^{e}} `The Book of Rays'.} fulfils this requirement. It
is a generalised multi-tape two-level system which is being used in
developing non-linear grammars.

This paper (1) presents the algorithms behind {\tt SemHe}; (2)
discusses the issues involved in compiling non-linear
descriptions; and (3) proposes extension/solutions to make writing
non-linear rules easier and more elegant.  The paper assumes knowledge
of multi-tape two-level morphology \cite{Kay:87,Kiraz:94Coling}.

\section{Linguistic Descriptions}

\begin{figure*}
   \centering
   \fbox{\begin{minipage}{\linewidth}
   \label{ex:ktab}
   \begin{tabbing}
      {}\=    tl\_alphabet({\tt 1}, {\tt [c$_1$,c$_2$,c$_3$,v,$\flat$]}). \ 
         tl\_alphabet({\tt 2}, {\tt [k,t,b,$\flat$]}).\
         tl\_alphabet({\tt 3}, {\tt [a,e,$\flat$]}). \=\% lexical alphabets
      \kill
      \> tl\_alphabet({\tt 0}, {\tt [k,t,b,a,e]}).\> \% surface alphabet\\
         tl\_alphabet({\tt 1}, {\tt [c$_1$,c$_2$,c$_3$,v,$\flat$]}). \ 
         tl\_alphabet({\tt 2}, {\tt [k,t,b,$\flat$]}).\
         tl\_alphabet({\tt 3}, {\tt [a,e,$\flat$]}). \% lexical alphabets\\
      \> tl\_set({\tt radical, [k,t,b]}).
         tl\_set({\tt vowel, [a,e]}).\ 
         tl\_set({\tt c1c3, [c$_1$,c$_3$]}).\> \% variable sets \\
      \> \tlruleintablong{R1}{[[],[],[]]}{[[$\flat$],[$\flat$],[$\flat$]]}{[[],[],[]]}{=>}
                      {[]}{[]}{[]}
                      {[]|[[],[],[]]} \\
      \> \tlruleintablong{R2}{[[],[],[]]}{[[P],[C],[]]}{[[],[],[]]}{=>}
                      {[]}{[C]}{[]}
                      {[c1c3(P),radical(C)]|[[],[],[]]}\\
      \> \tlruleintablong{R3}{[[],[],[]]}{[[v],[],[V]]}{[[],[],[]]}{=>}
                      {[]}{[V]}{[]}
                      {[vowel(V)]|[[],[],[]]}\\
      \> \tlruleintablong{R4}{[[],[],[]]}{[[v],[],[V]]}{[[c$_2$,v],[],[]]}{<=>}
                      {[]}{[]}{[]}
                      {[vowel(V)]|[[],[],[]]}\\
      \> \tlruleintablong{R5}{[[],[],[]]}{[[c$_2$],[C],[]]}{[[],[],[]]}{<=>}
                      {[]}{[C]}{[]}
                      {[radical(C)]|[[], [root:[measure=p`al]], []]}\\
      \> \tlruleintablong{R6}{[[],[],[]]}{[[c$_2$],[C],[]]}{[[],[],[]]}{<=>}
                      {[]}{[C,C]}{[]}
                      {[radical(C)]|[[], [root:[measure=pa``el]], []]}
   \end{tabbing}
   \end{minipage}}\\
   {\bf Listing 1}
\end{figure*}

The linguist provides {\tt SemHe} with three pieces of data: a
lexicon, two-level rules and word formation grammar. All entries take
the form of Prolog terms.\footnote{We describe here the terms which
are relevant to this paper. For a full description, see
\cite{Kiraz:thesis}.}  (Identifiers starting with an uppercase letter
denote variables, otherwise they are instantiated symbols.)  A lexical
entry is described by the term
\[ {\tt synword}(\langle morpheme\rangle, \langle category\rangle). \]
Categories are of the form
\begin{tabbing}
$\langle category\_symbol\rangle:[$\= $\langle feature\_attr_1=value_1\rangle,$\\
                                  \> $\ldots,$\\ 
                                  \> $\langle feature\_attr_n=value_n\rangle]$
\end{tabbing}
a notational variant of the PATR-II category formalism \cite{Shieber:86}.

A two-level rule is described using a syntactic variant of the
formalism described by \cite{Ruessink:89,Pulman:93}, including the
extensions by \cite{Kiraz:94Coling},

\begin{tabbing}
${\tt tl\_rule}(\langle id\rangle,$\= $\langle LLC\rangle,\langle Lex\rangle,
         \langle RLC\rangle,\langle Op\rangle,$\\
         \> $\langle LSC\rangle,\langle Surf\rangle, 
         \langle RSC\rangle,$\\
         \> $\langle variables\rangle,\langle features\rangle).$
\end{tabbing}
The arguments are: (1) a rule identifier, {\em id}; (2) the
left-lexical-context, {\em LLC}, the lexical center, {\em Lex}, and
the right-lexical-context, {\em RLC}, each in the form of a
list-of-lists, where the {\em i}th list represents the {\em i}th
lexical tape; (3) an operator, {\tt =>} for optional rules or {\tt
<=>} for obligatory rules; (4) the left-surface-context, {\em LSC},
the surface center, {\em Surf}, and the right-surface-context, {\em
RSC}, each in the form of a list; (5) a list of the {\em variables}
used in the lexical and surface expressions, each member in the form
of a predicate indicating the set identifier (see {\em infra}) and an
argument indicating the variable in question; and (6) a set of {\em
features} (i.e.~category forms) in the form of a list-of-lists, where
the {\em i}th item must unify with the feature-structure of the
morpheme affected by the rule on the {\em i}th lexical tape.  

A lexical string maps to a surface string iff (1) they can be
partitioned into pairs of lexical-surface subsequences, where each
pair is licenced by a rule, and (2) no partition violates an
obligatory rule.

Alphabet declarations take the form\break
{\tt tl\_alphabet}({\em\lab
tape\rab}, {\em\lab symbol\_list\rab}), and variable sets are described by the
predicate {\tt tl\_set}({\em\lab id\rab}, {\em\lab
symbol\_list\rab}). Word formation rules take the form of
unification-based CFG rules,
${\tt synrule}(\langle identifier\rangle,\break
                \langle mother\rangle,
                [\langle daughter_1\rangle,\ldots,
\langle daughter_n\rangle]).$


The following example illustrates the derivation of Syriac
/ktab/\footnote{Spirantization is ignored here; 
for a discussion on Syriac spirantization, 
see \cite{Kiraz:95RQ}.} `he wrote' (in the
simple {\em p`al} measure)\footnote{Syriac verbs are classified under
various measures (forms). The basic ones are: {\em p`al, pa``el} and
{\em 'af`el}.} from the pattern morpheme \{cvcvc\} `verbal pattern',
root \{ktb\} `notion of writing', and vocalism \{a\}. The three
morphemes produce the underlying form */katab/, which surfaces as
/ktab/ since short vowels in open unstressed syllables are
deleted. The process is illustrated in \exnum{+1}{}.\footnote{This
analysis is along the lines of \cite{McCarthy:81} -- based on
autosegmental phonology \cite{Goldsmith:76}.}
\begin{ex}
   \autoseg{k-t-b|cvcvc|-a---}{21x,23x,21x,23ll,21x} = */katab/ 
      $\Longrightarrow$
      /ktab/
\end{ex}
The {\em pa``el} measure of the same verb, viz.~/katteb/, is derived by
the gemination of the middle consonant (i.e.~t) and applying the 
appropriate vocalism \{ae\}.

The two-level grammar (Listing 1) assumes three lexical
tapes. Uninstantiated contexts are denoted by an empty list.  R1 is
the morpheme boundary (= $\flat$) rule. R2 and R3 sanction stem
consonants and vowels, respectively. R4 is the obligatory vowel
deletion rule. R5 and R6 map the second radical, [t], for {\em p`al}
and {\em pa``el} forms, respectively. In this example, the lexicon
contains the entries in \exnum{+1}{}.\footnote{Spreading is ignored here;
for a discussion, 
see \cite{Kiraz:94Coling}.}
\begin{ex}
$\mbox{synword}({\tt c_1vc_2vc_3, pattern:[ ]}).$\\
$\mbox{synword}({\tt ktb, root:[measure=M]}).$\\
$\mbox{synword}({\tt aa, vocalism:[measure=p`al]}).$\\
$\mbox{synword}({\tt ae, vocalism:[measure=pa``el]}).$
\end{ex}
Note that the value of `measure' in the root entry is uninstantiated; 
it is determined from the feature values in R5, R6 and/or the word grammar
(see {\em infra}, \S\ref{word-grammar}).

\section{Implementation}

There are two current methods for implementing two-level rules (both
implemented in {\tt SemHe}): (1)~compiling rules into finite-state
automata (multi-tape transducers in our case), and (2)~interpreting
rules directly. The former provides better performance, while the
latter facilitates the debugging of grammars (by tracing and by
providing debugging utilities along the lines of
\cite{Carter:95}). Additionally, the interpreter facilitates the
incremental compilation of rules by simply allowing the user to toggle
rules on and off.

The compilation of the above formalism into automata is described by
\cite{Grimley-Evans:96}. The following is a description
of the interpreter.

\subsection{Internal Representation}

The word grammar is compiled into a shift-reduce
parser. In addition, a first-and-follow algorithm, based on
\cite{Aho:77}, is applied to compute the feasible follow categories
for each category type. The set of feasible follow categories, {\em
NextCats}, of a particular category {\em Cat} is returned by the
predicate {\sc Follow}(+{\em Cat}, --{\em NextCats}).  Additionally, {\sc
Follow}({\tt bos}, {\em NextCats}) returns the set of category symbols
at the beginning of strings, and {\tt eos} $\in$ {\em NextCats}
indicates that {\em Cat} may occur at the end of strings.

The lexical component is implemented as character tries
\cite{Knuth:73}, one per tape. Given a list of lexical strings, {\em
Lex}, and a list of lexical pointers, {\em LexPtrs}, the predicate
\begin{tabbing}
$\mbox{{\sc Lexical-}}$\=$\mbox{{\sc Transitions}}(+Lex,\ +LexPtrs,$\\
                                   \> $-NewLexPtrs,\ -LexCats)$
\end{tabbing}
succeeds iff there are transitions on {\em Lex} from {\em
LexPtrs}; it returns {\em NewLexPtrs}, and the categories,
{\em LexCats}, at the end of morphemes, if any.

Two-level predicates are converted into an internal representation:
(1) every left-context expression is reversed and appended to an
uninstantiated tail; (2) every right-context expression is appended to
an uninstantiated tail; and (3) each rule is assigned a 6-bit
`precedence value' where every bit represents one of the six lexical
and surface expressions.  If an expression is {\em not} an empty list
(i.e.~context is specified), the relevant bit is set.  In analysis,
surface expressions are assigned the most significant bits, while
lexical expressions are assigned the least significant ones. In
generation, the opposite state of affairs holds. Rules are then
reasserted in the order of their precedence value. This ensures that
rules which contain the most specified expressions are tested first
resulting in better performance.

\subsection{The Interpreter Algorithm}

\begin{figure*}
   \centering
   \fbox{\begin{minipage}{\linewidth}
   \begin{tabbing}
   {\sc Partition}({\em SurfDone, SurfToDo, LexDone, LexToDo,
                       LexPtrs, NextCats, Result})\\
   mmm\=mmm\=\hspace{1.5in}\= \kill
   {\em SurfToDo} = {\tt []} {\bf \&} 
      \>\>\>\% surface string exhausted \\
   {\em LexToDo} =  {\tt [[],[],$\cdots$,[]]} {\bf \&} 
      \>\>\> \% all lexical strings exhausted \\
   {\em LexPtrs} = {\tt [rt,rt,$\cdots$,rt]} {\bf \&} 
      \>\>\> \% all lexical pointers are at the root node \\
   {\tt eos} $\in$ {\em NextCats}  {\bf \&}
      \>\>\> \% end-of-string\\
   {\em Result} = {\tt []}.
      \>\>\> \% output: no more results
   \end{tabbing}
   \end{minipage}}\\
   {\bf Listing 2}
\end{figure*}

\begin{figure*}
   \centering
   \fbox{\begin{minipage}{\linewidth}
   \begin{tabbing}
   {\sc Partition}(\= {\em SurfDone, SurfToDo, LexDone, LexToDo,}
                       {\em LexPtrs, NextCats},\\
                    \> [{\em ResultHead $\mid$ ResultTail}])\\
   mmm\=mmm\=\hspace{2.5in}\= \kill
   {\bf there is} 
         tl\_rule({\em Id, LLC, Lex, RLC, Op, LSC, Surf, RSC, Variables, 
                  Features}) {\bf such that}\\
   \> ({\em Op} = ({\tt =>} {\bf or} {\tt <=>}), {\em LexDone} = {\em LLC},
           {\em SurfDone} = {\em LSC},\\
   \>{\em SurfToDo} = {\em Surf} + {\em RSC} {\bf and}
           {\em LexToDo} = {\em Lex} + {\em RLC}) {\bf \&}\\
   {\sc Lexical-Transitions}%
                ({\em Lex, LexPtrs, NewLexPtrs, LexCats}) {\bf \&}\\

   {\bf push} {\em Features} {\bf onto} {\em FeatureStack} {\bf \&}
      \>\>\> \% keep track of rule features\\

   {\bf if} {\em LexCats} $\neq$ {\tt nil} {\bf then}
     \>\>\> \% found a morpheme boundary? \\

   \>{\bf while} {\em FeatureStack} {\bf is not empty}
      \>\> \% unify rule and lexical features\\
   \>\> {\bf unify} {\em LexCats} {\bf with} 
        ({\bf pop} {\em FeatureStack}) {\bf \&}\\

   \> {\bf push} {\em LexCats} {\bf onto} {\em ParseStack} {\bf \&}
      \>\> \% update the parse stack\\

   \> {\bf if} {\em LexCats} $\in$ {\em NextCats} {\bf then}
      \>\> \% get next category\\
   \>\>   {\sc Follow}({\em LexCats, NewNextCats})\\
   {\bf end if \&}\\

   {\em ResultHead} = \={\em Id/SurfDone/Surf/RSC/}\\
   \> {\em LexDone/Lex/RLC/LexCats} {\bf \&}\\

   mmm\=mmm\=\hspace{2.5in}\= \kill
   {\em NewSurfDone} = {\em SurfDone} + {\bf reverse} {\em Surf} 
                {\bf \&} \>\>\>  \% make new arguments ... \\
   {\em NewSurfToDo} = {\em RSC} {\bf \&}
      \>\>\>  \% ... and recurse\\
   {\em NewLexDone} = {\em LexDone} + {\bf reverse} {\em Lex} 
                {\bf \&}\\
   {\em NewLexToDo} = {\em RLC} {\bf \&}\\
      {\sc Partition}(\= {\em NewSurfDone, NewSurfToDo},\\
                  \> \> {\em NewLexDone, 
                           NewLexToDo,}\\
                  \> \> {\em NewLexPtrs, NewNextCats,
                         ResultTail}) {\bf \&}\\

   mmm\=mmm\=\hspace{2.5in}\= \kill
   {\bf for all} {\em SetId(Var)} $\in$ {\em Variables}
      \>\>\> \% check variables\\
   \>{\bf there is} tl\_set({\em SetId, Set}) {\bf such that}
                 {\em Var} $\in$ {\em Set}.
   \end{tabbing}
   \end{minipage}}\\
   {\bf Listing 3}
\end{figure*}

\begin{figure*}
   \centering
   \fbox{\begin{minipage}{\linewidth}
   \begin{tabbing}
   {\sc Coerce}(\=[{\em Id/LSC/Surf/RSC/LLC/Lex/RLC/LexCats} $\mid$ 
      {\em ResultTail}], {\em PrevCats}, \\
      \> [{\em Id/Surf/Lex} $\mid$ {\em PartitionTail}])\\
   \=mmm\=mmm\kill
   {\bf if} {\em LexCats} $\neq$ {\tt nil} {\bf then} \\
      \>\> {\em CurrentCats} = {\em LexCats}\\
      {\bf else}\\ \>\> {\em CurrentCats} = {\em PrevCats} {\bf \&}\\

   {\bf not} 
      {\sc Invalid-Partition}({\em LSC, Surf, RSC, LLC, Lex, RLC, CurrentCats})
      {\bf \&}\\
      {\sc Coerce}({\em ResultTail, CurrentCats, PartitionTail}).
   \end{tabbing}
  \end{minipage}}\\
   {\bf Listing 4}
\end{figure*}

\begin{figure*}
   \centering
   \fbox{\begin{minipage}{\linewidth}
   \begin{tabbing}
   {\sc Invalid-Partition}({\em LSC, Surf, RSC, LLC, Lex, RLC, Cats})\\
   mmm\=mmm\=\hspace{2in}\= \kill
   {\bf there is} 
         tl\_rule({\em Id, LLC, Lex, RLC, {\tt <=>}, LSC, NotSurf, RSC, 
                  Variables, Features}) {\bf such that} \\
   \>{\em NotSurf} $\neq$ {\em Surf} {\bf \&}\\

   {\bf for all} {\em SetId(Var)} $\in$ {\em Variables}
      \>\>\>  \% check variables\\
   \>{\bf there is} tl\_set({\em SetId, Set}) {\bf such that}
                   {\em Var} $\in$ {\em Set} {\bf \&}\\
   {\bf unify} {\em Cats} {\bf with} {\em Features} {\bf \&}\\
   {\bf fail}.
   \end{tabbing}
   \end{minipage}}\\
   {\bf Listing 5}
\end{figure*}

\begin{figure*}
   \centering
   \fbox{\begin{minipage}{\linewidth}
   \begin{tabbing}
   {\sc Two-Level-Analysis}(?{\em Surf}, ?{\em Lex}, -{\em Partition}, -{\em Parse}) \\
   \=mmm\= \kill
      {\sc Follow}({\tt bos}, {\em NextCats}) {\bf \&}\\
      {\sc Partition}(\={\tt []}, {\em Surf}, 
                       {\tt [[],[],$\cdots$,[]]}, {\em Lex},
                       {\tt [rt,rt,$\cdots$,rt]}, {\em NextCats},
                       {\em Result}) {\bf \&} \\
      {\sc Coerce}({\bf reverse} {\em Result}, {\tt nil}, 
           {\em Partition}) {\bf \&} \\
      {\sc Shift-Reduce}({\em ParseStack}, {\em Parse}).
   \end{tabbing}
   \end{minipage}}\\
   {\bf Listing 6}
\end{figure*}

The algorithms presented below are given in terms of prolog-like
non-deterministic operations. A clause is satisfied iff all the
conditions under it are satisfied.
The predicates are depicted top-down in \exnum{+1}{}. 
({\tt SemHe} makes use of an earlier implementation by \cite{Pulman:93}.)

\begin{ex} \ \\
   \psfig{figure=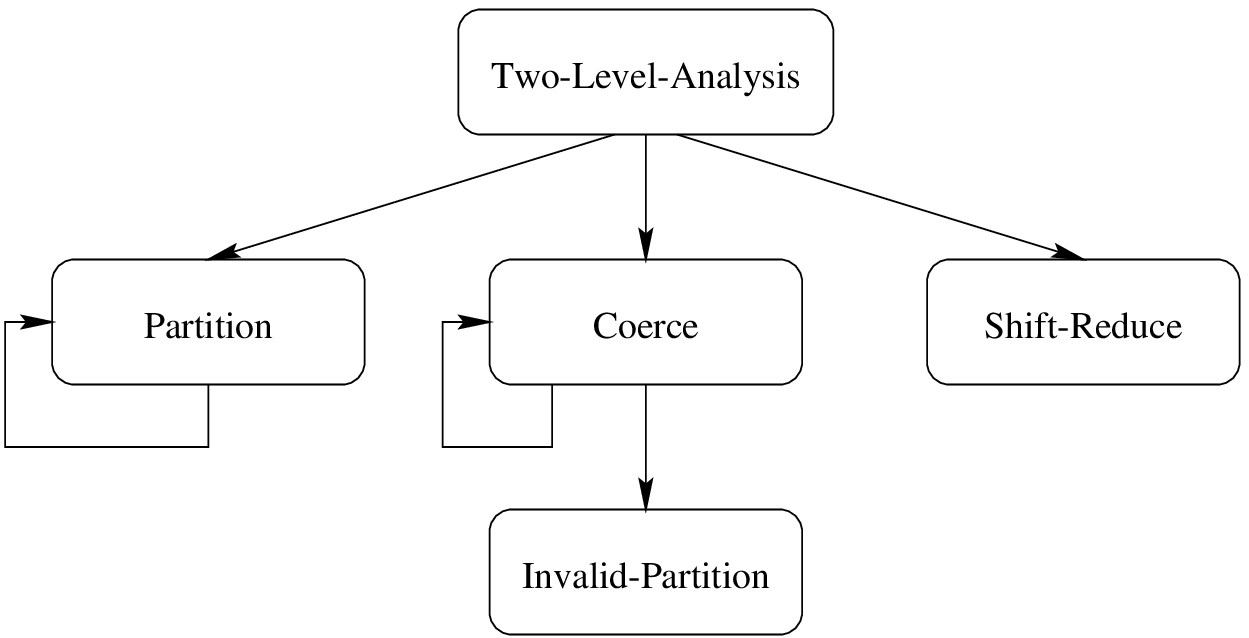,width=2.75in}
\end{ex}

In order to minimise accumulator-passing arguments, we assume the
following initially-empty stacks: {\em ParseStack} accumulates the
category structures of the morphemes identified, and {\em
FeatureStack} maintains the rule features encountered so far.  
(`+' indicates concatenation.)

{\sc Partition} partitions a two-level analysis into sequences of
lexical-surface pairs, each licenced by a rule. The base case of the
predicate is given in Listing 2,\footnote{For efficiency, variables
appearing in left-context and centre expressions are evaluated after
{\sc Lexical-Transitions} since they will be fully instantiated then;
only right-contexts are evaluated after the recursion.} and
the recursive case in Listing 3.

The recursive {\sc Coerce} predicate ensures that no partition is
violated by an obligatory rule.  It takes three arguments: {\em
Result} is the output of {\sc Partition} (usually reversed by the
calling predicate, hence, {\sc Coerce} deals with the last partition
first), {\em PrevCats} is a register which keeps track of the last
morpheme category encountered, and {\em Partition} returns selected
elements from {\em Result}. The base case of the predicate is simply
{\sc Coerce}({\tt [], \_, []}) -- i.e., no more partitions.  
The recursive case is shown in Listing 4. {\em CurrentCats} keeps
track of the category of the morpheme which occures in the current
partition.  The invalidity of a partition is determined by {\sc
Invalid-Partition} (Listing 5).

{\sc Two-Level-Analysis} (Listing 6) is the main predicate. It takes a
surface string or lexical string(s) and returns a list of partitions
and a morphosyntactic parse tree.  To analyse a surface form, one
calls {\sc Two-Level-Analysis}(+{\em Surf}, --{\em Lex}, --{\em
Partition}, --{\em Parse}).  To generate a surface form, one calls
{\sc Two-Level-Analysis}(--{\em Surf}, +{\em Lex}, --{\em Partition},
--{\em Parse}).

  \section{Developing Non-Linear Grammars}
  \label{testing}

\begin{figure*}
   \centering
   \fbox{\begin{minipage}{\linewidth}
   \begin{tabbing}
      {}\= tl\_set({\tt radical, [k,t,b]}).\kill
      \> \tlruleintablong{R7}{[[],[],[]]}{[[v],[],[V]]}{[[c$_3$,$\flat$,e],[],[]]}{<=>}
                      {[]}{[]}{[]}
                      {[vowel(V)]|[[],[],[]]}\\
      \> \tlruleintablong{R8}{[]}{[V1]}{[C,V2]}{<=>}
                      {[]}{[]}{[]}
                      {[vowel(V1),vowel(V2),radical(C)]|[[],[],[]]}
   \end{tabbing}
   \end{minipage}}\\
   {\bf Listing 7}
\end{figure*}

When developing Semitic grammars, one comes across various issues and
problems which normally do not arise with linear grammars. Some can be
solved by known methods or `tricks'; others require extensions in
order to make developing grammars easier and more elegant.
  This
section discuss issues which normally do not arise when compiling
linear grammars.

 \subsection{Linearity vs. Non-Linearity}

In Semitic languages, non-linearity occurs only in stems. Hence,
lexical descriptions of stems make use of three lexical tapes
(pattern, root \& vocalism), while those of prefixes and suffixes use
the first lexical tape. This requires duplicating rules when stating
lexical constraints. Consider rule R4 (Listing 1). It allows the
deletion of the first stem vowel by the virtue of $RLC$ (even if ${\tt
c_2}$ was not indexed); hence /katab/ $\rightarrow$ /ktab/. Now
consider adding the suffix \{eh\} `him/it': /katab/+\{eh\}
$\rightarrow$ /katbeh/, where the second stem vowel is deleted since
deletion applies right-to-left; however, $RLC$ can only cope with stem
vowels. Rule R7 (Listing 7) is required. One might suggest
placing constraints on surface expressions instead. However, doing so
causes surface expressions to be dependent on other rules.

Additionally, $Lex$ in R4 and R7 deletes stem
vowels. Consider adding the prefix \{wa\} `and': \{wa\} + /katab/ +
\{eh\} $\rightarrow$ /wkatbeh/, where the prefix vowel is also
deleted. To cope with this, two additional rules like R4 and R7 are
required, but with $Lex$ = {\tt [[V],[],[]]}.  

We resolve this by allowing the user to write expansion rules of the
from
\[ {\tt expand}(\langle symbol\rangle, \langle expansion\rangle,
                      \langle variables\rangle). \]
In our example, the expansion rules in \exnum{+1}{} are needed.
\begin{ex}
   expand({\tt C}, {\tt [[C],[],[]]}, {\tt [radical(C)]}).\\
   expand({\tt C}, {\tt [[c],[C],[]]}, {\tt [radical(C)]}).\\
   expand({\tt V}, {\tt [[V],[],[]]}, {\tt [vowel(V)]}).\\
   expand({\tt V}, {\tt [[v],[],[V]]}, {\tt [vowel(V)]}).
\end{ex}
The linguist can then rewrite R4 as R8 (Listing 7), and expand it with
the command {\tt expand(R8)}. This produces four rules of the form of
R4, but with the following expressions for $Lex$ and
$RLC$:\footnote{Note, however, that the {\tt expand} command does not
insert $\flat$ randomly in context expressions.}

\begin{tabular}{ll}
   $Lex$ & $RLC$ \\
   {\tt [[V1],[],[]]} & {\tt [[C,V2],[],[]]} \\
   {\tt [[V1],[],[]]} & {\tt [[c,v],[C],[V2]]} \\
   {\tt [[v],[],[V1]]} & {\tt [[C,V2],[],[]]} \\
   {\tt [[v],[],[V1]]} & {\tt [[c,v],[C],[V2]]}
\end{tabular}

 \subsection{Vocalisation}

Orthographically, Semitic texts are written without short vowels.  It
was suggested by \cite[et.~seq.]{Beesley:89} and \cite{Kiraz:94Coling}
to allow short vowels to be optionally deleted. This, however, puts a
constraint on the grammar: no surface expression can contain a vowel,
lest the vowel is optionally deleted.

We assume full vocalisation in writing rules. A second set of rules
can allow the deletion of vowels. The whole grammar can be taken as
the composition of the two grammars: e.g. \{cvcvc\},\{ktb\},\{aa\}
$\rightarrow$ /ktab/ $\rightarrow$ [ktab, ktb].

 \subsection{Morphosyntactic Issues}
\label{word-grammar}
\begin{figure*}
   \centering
   \fbox{\begin{minipage}{\linewidth}
   \begin{tabbing}
      synrule({\tt rule1}, \= {\tt stem:[\Xbar=-2,measure=M,measure=p`al|pa``el]},\\
                           \> [\={\tt pattern:[]},
                           {\tt root:[measure=M,measure=p`al|pa``el]}, \\
                           \> \> {\tt vocalism:[measure=M,measure=p`al|pa``el]}]).\\
      synrule({\tt rule2}, \= {\tt stem:[\Xbar=-2,measure=M]},\\
                           \> [\={\tt stem\_affix:[measure=M]}, \\
                           \>\>  {\tt pattern:[]},
                             {\tt root:[measure=M]},
                             {\tt vocalism:[measure=M]}]).\\
      synrule({\tt rule3}, \= {\tt stem:[\Xbar=-1,measure=M,mood=act]},\\
                           \> [\={\tt stem:[bar=-2,measure=M,mood=act]}]).\\
      synrule({\tt rule4}, \= {\tt stem:[\Xbar=-1,measure=M,mood=pass]},\\
                           \> [\={\tt reflexive:[]},
                              {\tt stem:[\Xbar=-2,measure=M,mood=pass]}]).\\
      synrule({\tt rule5}, \= {\tt stem:[\Xbar=0,measure=M,mood=MD,npg=s\&3\&m]},\\
                           \> [\={\tt stem:[\Xbar=-1,measure=M,mood=MD]}]).\\
      synrule({\tt rule6}, \= {\tt stem:[\Xbar=0,measure=M,mood=MD,npg=NPG]},\\
                           \> [\={\tt stem:[\Xbar=-1,measure=M,mood=MD]},
                               {\tt vim:[type=suff,circum=no,npg=NPG]}]).\\
      synrule({\tt rule7}, \= {\tt stem:[\Xbar=0,measure=M,mood=MD,npg=NPG]},\\
                           \> [\={\tt vim:[type=pref,circum=no,npg=NPG]},
                               {\tt stem:[\Xbar=-1,measure=M,mood=MD]}]).\\
      synrule({\tt rule8}, \= {\tt stem:[\Xbar=0,measure=M,mood=MD,npg=NPG]},\\
                           \> [\={\tt vim:[type=pref,circum=yes,npg=NPG]},
                              {\tt stem:[\Xbar=-1,measure=M,mood=MD]},\\
                           \>\> {\tt vim:[type=suff,circum=yes,npg=NPG]}]).
   \end{tabbing}
  \end{minipage}}\\
   {\bf Listing 8}
\end{figure*}


Finite-state models of two-level morphology implement morphotactics in
two ways: using `continuation patterns/classes'
\cite{Koskenniemi:83,Antworth:90,Karttunen:93} or unification-based
grammars
\cite{Bear:86,Ritchie:92book}.  The former fails to provide elegant
morphosyntactic parsing for Semitic languages, as will be illustrated
in this section.

  \subsubsection{Stems and \Xbar-Theory}

A pattern, a root and a vocalism do not alway produce a free stem
which can stand on its own. In Syriac, for example, some verbal forms
are bound: they require a {\bf stem morpheme} which indicates the
measure in question, e.g.~the prefix \{\A a\} for {\em af`el}
stems. Additionally, passive forms are marked by the {\bf reflexive
morpheme} \{\A et\}, while active forms are not marked at all.

This structure of stems can be handled hierarchically using
\Xbar-theory.  A stem whose stem morpheme is known is assigned {\tt
\Xbar=-2} (Rules 1-2 in Listing 8).  Rules which indicate mood can
apply only to stems whose measure has been identified (i.e.~they have
{\tt \Xbar=-2}). The resulting stems are assigned {\tt \Xbar=-1}
(Rules 3-4 in Listing 8). The parsing of Syriac /\A etkteb/ (from \{\A
et\}+/kateb/ after the deletion of /a/ by R4) appears in
\exnum{+1}{}.\footnote{In the remaining examples, it is assumed that
the lexicon and two-level rules are expanded to cater for the new
material.}
\begin{ex}
   \begin{parsetree}
     (.stem:[\Xbar=-1]. (.reflexive. .\A et.)
                      (.stem:[\Xbar=-2]. (.pattern. .cvcvc.)
                                       (.root. .ktb.)
                                       (.vocalism. .ae.)
                      )
     )
   \end{parsetree}
\end{ex}

Now free stems which may stand on their own can be assigned {\tt
\Xbar=0}. However, some stems require verbal inflectional markers.

  \subsubsection{Verbal Inflectional Markers}
   \label{more-circum}

\begin{figure*}
   \centering
   \fbox{\begin{minipage}{5.5in}
\begin{tabular}{lllll}
   Verb Class & Inflections Analysed & 1st Analysis& Subsequent Analysis & Mean\\
              &                      & (sec/word)  & (sec/word) & (sec/word)\\
   \hline
   Strong                  & 78 & 5.053   & 0.028  & 2.539 \\
   Initial {\em n\={u}n}   & 52 & 6.756   & 0.048  & 3.404 \\
   Initial {\em \={a}laph} & 57 & 4.379   & 0.077  & 2.228\\
   Middle {\em \={a}laph}  & 67 & 5.107   & 0.061  & 2.584 \\
   Overall mean            & 63.5& 5.324   & 0.054  & {\bf 2.689}
\end{tabular}
   \end{minipage}}\\
   {\bf Table 1}
\end{figure*}

With respect to verbal inflexional markers (VIMs), there are various
types of Semitic verbs: those which do not require a VIM
(e.g.~sing.~3rd masc.), and those which require a VIM in the form of a
prefix (e.g.~perfect), suffix (e.g.~some imperfect forms), or
circumfix (e.g.~other imperfect forms).

Each VIM is lexically marked {\em inter alia} with two features: `type'
which states whether it is a prefix or a suffix, and `circum' which
denotes whether it is a circumfix.
Rules 5-8 (Listing 8) handle this.

The parsing of Syriac /netkatbun/ (from \{ne\}+ \{\A
et\}+/katab/+\{un\}) appears in \exnum{+1}{}.
\begin{ex}
   \begin{parsetree}
      (.stem:[\Xbar=0].
        (.vim. .ne.)
        (.stem:[\Xbar=-1]. (.reflexive. .\A et.)
                         (.stem:[\Xbar=-2]. (.pattern. .cvcvc.)
                                          (.root. .ktb.)
                                          (.vocalism. .aa.)
                         )
        )
        (.vim. .un.)
      )
   \end{parsetree}
\end{ex}

{}\cite{Beesley:89} handle this problem by finding a logical
expression for the prefix and suffix portions of circumfix morphemes,
and use unification to generate only the correct forms -- see
\cite[p.~158]{Sproat:92}. This approach, however, cannot be used here
since, unlike Arabic, not all Syriac VIMs are in the form of
circumfixes.

  \subsubsection{Interfacing with a Syntactic Parser}

A Semitic `word' (string separated by word boundary) may in fact be a
clause or a sentence. Therefore, a morphosyntactic parsing of a `word'
may be a (partial) syntactic parsing of a sentence in the form of a
(partial) tree. The output of a
morphological analyser can be structured in a manner suitable for
syntactic processing. Using tree-adjoining grammars \cite{Joshi:85}
might be a possibility.

 \section{Performance}

To test the integrity, robustness and performance of the
implementation, a two-level grammar of the most frequent words in the
Syriac New Testament was compiled based on the data in
\cite{Kiraz:94LexTools}. The grammar covers most classes of
verbal and nominal forms, in addition to prepositions, proper nouns
and words of Greek origin. A wider coverage would involve enlarging
the lexicon (currently there are 165 entries) and might triple the
number of two-level rules (currently there are {\em c.}~50 rules).

Table 1 provides the results of analysing verbal classes. The test for
each class represents analysing most of its inflexions. The test was
executed on a Sparc ELC computer.

By constructing a corpus which consists only of the most frequent
words, one can estimate the performance of analysing the corpus as
follows,

\[ P = \frac{5.324n + \sum_{i=1}^{n} 0.054(f_i-1)}{\sum_{i=1}^{n} f_i} \ \ \mbox{sec/word} \]
where $n$ is the number of distinct words in the corpus and $f_i$ is
the frequency of occurrence of the $i$th word. The SEDRA database
\cite{Kiraz:94OCA} provides such data. All occurrences of the 100 most
frequent lexemes in their various inflections (a total of 72,240
occurrences) can be analysed at the rate of 16.35
words/sec. (Performance will be less if additional rules are added for
larger coverage.)

The results may not seem satisfactory when compared with other prolog
implementations of the same formalism (cf.~50 words/sec,
in \cite{Carter:95}). One should, however, keep in mind the complexity of
Syriac morphology. In addition to morphological non-linearity,
phonological conditional changes -- consonantal and vocalic -- occur
in all stems, and it is not unusual to have more than five such
changes per word. Once developed, a grammar is usually compiled
into automata which provides better performance.

\section{Conclusion}

This paper has presented a computational morphology system which is
adequate for handling non-linear grammars. We are currently expanding
the grammar to cover the whole of New Testament Syriac. One of our
future goals is to optimise the prolog implementation for speedy
processing and to add debugging facilities along the lines of
\cite{Carter:95}.

For useful results, a Semitic morphological analyser needs to interact
with a syntactic parser in order to resolve ambiguities. Most
non-vocalised strings give more than one solution, and some
inflectional forms are homographs even if fully vocalised (e.g.~in
Syriac imperfect verbs: sing.~3rd masc. = plural 1st common, and
sing.~3rd fem. = sing.~2nd masc.). We mentioned earlier the possibility
of using TAGs.


\begin{thebibliography}{fullname}

\bibitem[Aho and Ullman, 1977]{Aho:77}
Aho, A. and Ullman, J. (1977).
\newblock {\em Principles of Compiler Design}.
\newblock Addison-Wesley.

\bibitem[Antworth, 1990]{Antworth:90}
Antworth, E. (1990).
\newblock {\em PC-KIMMO: A two-Level Processor for Morphological Analysis}.
\newblock Occasional Publications in Academic Computing 16. Summer Institute of
  Linguistics, Dallas.

\bibitem[Bear, 1986]{Bear:86}
Bear, J. (1986).
\newblock A morphological recognizer with syntactic and phonological rules.
\newblock In {\em COLING-86}, pages 272--6.

\bibitem[Beesley, 1990]{Beesley:90}
Beesley, K. (1990).
\newblock Finite-state description of {Arabic} morphology.
\newblock In {\em Proceedings of the Second Cambridge Conference: Bilingual
  Computing in Arabic and English}.

\bibitem[Beesley, 1991]{Beesley:91}
Beesley, K. (1991).
\newblock Computer analysis of {Arabic} morphology.
\newblock In Comrie, B. and Eid, M., editors, {\em Perspectives on Arabic
  Linguistics III: Papers from the Third Annual Symposium on Arabic
  Linguistics}. Benjamins, Amsterdam.

\bibitem[Beesley et~al., 1989]{Beesley:89}
Beesley, K., Buckwalter, T., and Newton, S. (1989).
\newblock Two-level finite-state analysis of {Arabic} morphology.
\newblock In {\em Proceedings of the Seminar on Bilingual Computing in Arabic
  and English}. The Literary and Linguistic Computing Centre, Cambridge.

\bibitem[Bird and Ellison, 1994]{Bird:94}
Bird, S. and Ellison, T. (1994).
\newblock One-level phonology.
\newblock {\em Computational Linguistics}, 20(1):55--90.

\bibitem[Carter, 1995]{Carter:95}
Carter, D. (1995).
\newblock Rapid development of morphological descriptions for full language
  processing systems.
\newblock In {\em EACL-95}, pages 202--9.

\bibitem[Goldsmith, 1976]{Goldsmith:76}
Goldsmith, J. (1976).
\newblock {\em Autosegmental Phonology}.
\newblock PhD thesis, MIT.
\newblock Published as {\em Autosegmental and Metrical Phonology}, Oxford 1990.

\bibitem[Grimley-Evans et~al., 1996]{Grimley-Evans:96}
Grimley-Evans, E., Kiraz, G., and Pulman, S. (1996).
\newblock Compiling a partition-based two-level formalism.
\newblock In {\em COLING-96}.
\newblock Forthcoming.

\bibitem[Joshi, 1985]{Joshi:85}
Joshi, A. (1985).
\newblock Tree-adjoining grammars: {How} much context sensitivity is required
  to provide reasonable structural descriptions.
\newblock In Dowty, D., Karttunen, L., and Zwicky, A., editors, {\em Natural
  Language Parsing}. Cambridge University Press.

\bibitem[Karttunen, 1983]{Karttunen:83}
Karttunen, L. (1983).
\newblock Kimmo: A general morphological processor.
\newblock {\em Texas Linguistic Forum}, 22:165--86.

\bibitem[Karttunen, 1993]{Karttunen:93}
Karttunen, L. (1993).
\newblock Finite-state lexicon compiler.
\newblock Technical report, Palo Alto Research Center, Xerox Corporation.

\bibitem[Karttunen and Beesley, 1992]{Karttunen:92}
Karttunen, L. and Beesley, K. (1992).
\newblock Two-level rule compiler.
\newblock Technical report, Palo Alto Research Center, Xerox Corporation.

\bibitem[Kataja and Koskenniemi, 1988]{Kataja:88}
Kataja, L. and Koskenniemi, K. (1988).
\newblock Finite state description of {Semitic} morphology.
\newblock In {\em COLING-88}, volume~1, pages 313--15.

\bibitem[Kay, 1987]{Kay:87}
Kay, M. (1987).
\newblock Nonconcatenative finite-state morphology.
\newblock In {\em EACL-87}, pages 2--10.

\bibitem[Kiraz, 1994a]{Kiraz:94OCA}
Kiraz, G. (1994a).
\newblock Automatic concordance generation of {Syriac} texts.
\newblock In Lavenant, R., editor, {\em VI Symposium Syriacum 1992}, Orientalia
  Christiana Analecta 247, pages 461--75. Pontificio Institutum Studiorum
  Orientalium.

\bibitem[Kiraz, 1994b]{Kiraz:94LexTools}
Kiraz, G. (1994b).
\newblock {\em Lexical Tools to the Syriac New Testament}.
\newblock JSOT Manuals 7. Sheffield Academic Press.

\bibitem[Kiraz, 1994c]{Kiraz:94Coling}
Kiraz, G. (1994c).
\newblock Multi-tape two-level morphology: a case study in {Semitic} non-linear
  morphology.
\newblock In {\em COLING-94}, volume~1, pages 180--6.

\bibitem[Kiraz, 1995]{Kiraz:95RQ}
Kiraz, G. (1995).
\newblock {\em Introduction to Syriac Spirantization}.
\newblock Bar Hebraeus Verlag, The Netherlands.

\bibitem[Kiraz, 1996]{Kiraz:thesis}
Kiraz, G. (1996).
\newblock {\em Computational Approach to Non-Linear Morphology}.
\newblock PhD thesis, University of Cambridge.

\bibitem[Knuth, 1973]{Knuth:73}
Knuth, D. (1973).
\newblock {\em The Art of Computer Programming}, volume~3.
\newblock Addison-Wesley.

\bibitem[Kornai, 1991]{Kornai:91}
Kornai, A. (1991).
\newblock {\em Formal Phonology}.
\newblock PhD thesis, Stanford University.

\bibitem[Koskenniemi, 1983]{Koskenniemi:83}
Koskenniemi, K. (1983).
\newblock {\em Two-Level Morphology}.
\newblock PhD thesis, University of Helsinki.

\bibitem[Lavie et~al., 1990]{Lavie:90}
Lavie, A., Itai, A., and Ornan, U. (1990).
\newblock On the applicability of two level morphology to the inflection of
  {Hebrew} verbs.
\newblock In Choueka, Y., editor, {\em Literary and Linguistic Computing 1988:
  Proceedings of the 15th International Conference}, pages 246--60.

\bibitem[McCarthy, 1981]{McCarthy:81}
McCarthy, J. (1981).
\newblock A prosodic theory of nonconcatenative morphology.
\newblock {\em Linguistic Inquiry}, 12(3):373--418.

\bibitem[Narayanan and Hashem, 1993]{Narayanan:93}
Narayanan, A. and Hashem, L. (1993).
\newblock On abstract finite-state morphology.
\newblock In {\em EACL-93}, pages 297--304.

\bibitem[Pulman and Hepple, 1993]{Pulman:93}
Pulman, S. and Hepple, M. (1993).
\newblock A feature-based formalism for two-level phonology: a description and
  implementation.
\newblock {\em Computer Speech and Language}, 7:333--58.

\bibitem[Ritchie et~al., 1992]{Ritchie:92book}
Ritchie, G., Black, A., Russell, G., and Pulman, S. (1992).
\newblock {\em Computational Morphology: Practical Mechanisms for the English
  Lexicon}.
\newblock MIT Press, Cambridge Mass.

\bibitem[Ruessink, 1989]{Ruessink:89}
Ruessink, H. (1989).
\newblock Two level formalisms.
\newblock Technical Report~5, Utrecht Working Papers in NLP.

\bibitem[Shieber, 1986]{Shieber:86}
Shieber, S. (1986).
\newblock {\em An Introduction to Unification-Based Approaches to Grammar}.
\newblock CSLI Lecture Notes Number 4. Center for the Study of Language and
  Information, Stanford.

\bibitem[Sproat, 1992]{Sproat:92}
Sproat, R. (1992).
\newblock {\em Morphology and Computation}.
\newblock MIT Press, Cambridge Mass.

\bibitem[Wiebe, 1992]{Wiebe:92}
Wiebe, B. (1992).
\newblock Modelling autosegmental phonology with multi-tape finite state
  transducers.
\newblock Master's thesis, Simon Fraser University.

\end{thebibliography}

\end{document}